\documentclass[Afour,sageh,times]{sagej}

\usepackage{moreverb,url}

\usepackage[colorlinks,bookmarksopen,bookmarksnumbered,citecolor=red,urlcolor=red]{hyperref}

\newcommand\BibTeX{{\rmfamily B\kern-.05em \textsc{i\kern-.025em b}\kern-.08em
T\kern-.1667em\lower.7ex\hbox{E}\kern-.125emX}}

\def\volumeyear{2025}

\begin{document}

\runninghead{Li et al.}

\title{Tactile SoftHand-A: 3D-Printed, Tactile, Highly-underactuated, Anthropomorphic Robot Hand with an Antagonistic Tendon Mechanism}

\author{Haoran Li\affilnum{1,}\affilnum{2}, Christopher J. Ford\affilnum{1}, Chenghua Lu\affilnum{1}, Yijiong Lin\affilnum{1}, Matteo Bianchi\affilnum{3},\\ Manuel G. Catalano\affilnum{4}, Efi Psomopoulou*\affilnum{1} and Nathan F. Lepora*\affilnum{1}}

\affiliation{\affilnum{1} School of Engineering Mathematics and Technology, and Bristol Robotics Laboratory, University of Bristol, Bristol, U.K.\\
\affilnum{2} School of Robotics, Xi'an Jiaotong-Liverpool University, China.\\
\affilnum{3} Department of Information Engineering and the Research Center "E.Piaggio", University of Pisa, Italy.\\
\affilnum{4} Soft Robotics for Human Cooperation and Rehabilitation, Istituto Italiano di Tecnologia (IIT), Italy.\\
\affilnum{*}These authors contributed equally to supervising this research.
}

\corrauth{Nathan F. Lepora, School of Engineering Mathematics and Technology, and Bristol Robotics Laboratory, University of Bristol, Bristol, U.K.}

\email{n.lepora@bristol.ac.uk}

\begin{abstract}   
A challenging and important problem for tendon-driven multi-fingered robotic hands is to ensure grasping adaptivity while minimizing the number of actuators needed to provide human-like functionality. Inspired by the Pisa/IIT SoftHand, this paper introduces a 3D-printed, highly-underactuated, tactile-sensorized, five-finger robotic hand named the Tactile SoftHand-A, which features an antagonistic mechanism to actively open and close the hand. Our proposed dual-tendon design gives options that allow active control of specific (distal or proximal interphalangeal) joints; for example, to adjust from an enclosing to fingertip grasp or to manipulate an object with a fingertip. We also develop and integrate a new design of fully 3D-printed vision-based tactile sensor within the fingers that requires minimal hand assembly. A control scheme based on analytically extracting contact location and slip from the tactile images is used to coordinate the antagonistic tendon mechanism (using a marker displacement density map, suitable for TacTip-based sensors). We perform extensive testing of a single finger, the entire hand, and the tactile capabilities to show the improvements in reactivity, load-bearing, and manipulability in comparison to a SoftHand that lacks the antagonistic mechanism. We also demonstrate the hand's reactivity to contact disturbances including slip, and how this enables teleoperated control from human hand gestures. Overall, this study points the way towards a class of low-cost, accessible, 3D-printable, tactile, underactuated human-like robotic hands, and we openly release the designs to facilitate others to build upon this work. The designs are open-sourced at \href{https://github.com/HaoranLi-Data/Tactile_SoftHand_A}{github.com/HaoranLi-Data/Tactile\_SoftHand\_A}
\end{abstract}


\maketitle

\section{Introduction}\label{intro}
Anthropomorphic robotic hands can be used in various ways, such as for grasping/pick-and-place, remote teleoperation, autonomous in-hand manipulation, and as prosthetics. However, developing economical and accessible anthropomorphic robotic hands that can approach the tactile sensibility and rich functionality of human hands has always been one of the most difficult and potentially rewarding challenges in robotics. The use of underactuated mechanisms is particularly appealing due to their simple structure, ease of maintenance, low cost, straightforward control, and adaptability to conform to the shape of objects~(\cite{piazza2019century,6663692}). To achieve this underactuation, a range of mechanisms have been proposed, including pneumatic~(\cite{9854144}, \cite{lu2024dexitac}), tendon-driven~(\cite{chen2018topology,catalano2014adaptive,li2022brl,mizushima2018multi}), and linkage-driven~(\cite{9137654}).

The article ``A Century of Robotic Hands''~(\cite{piazza2019century}) made some key observations about past trends and future directions in artificial hand design. One trend was towards a principled simplification of robotic hand design that maintains the advantages of an anthropomorphic structure while sensibly reducing the complexity of actuation and control. A second trend was an increased adoption of soft-robotics approaches, such as underactuated hands that utilize compliance or soft structures for adaptivity during grasping and manipulation. Both of these trends tend towards more biomimetic artificial hands, as human hands balance simplicity with functionality by making extensive use of compliance and underactuation.  

\begin{figure*}[t!]
	\centering
	\begin{tabular}[b]{c}
		\hspace{0cm}\includegraphics[width=\textwidth,trim={0 0 0 0},clip]{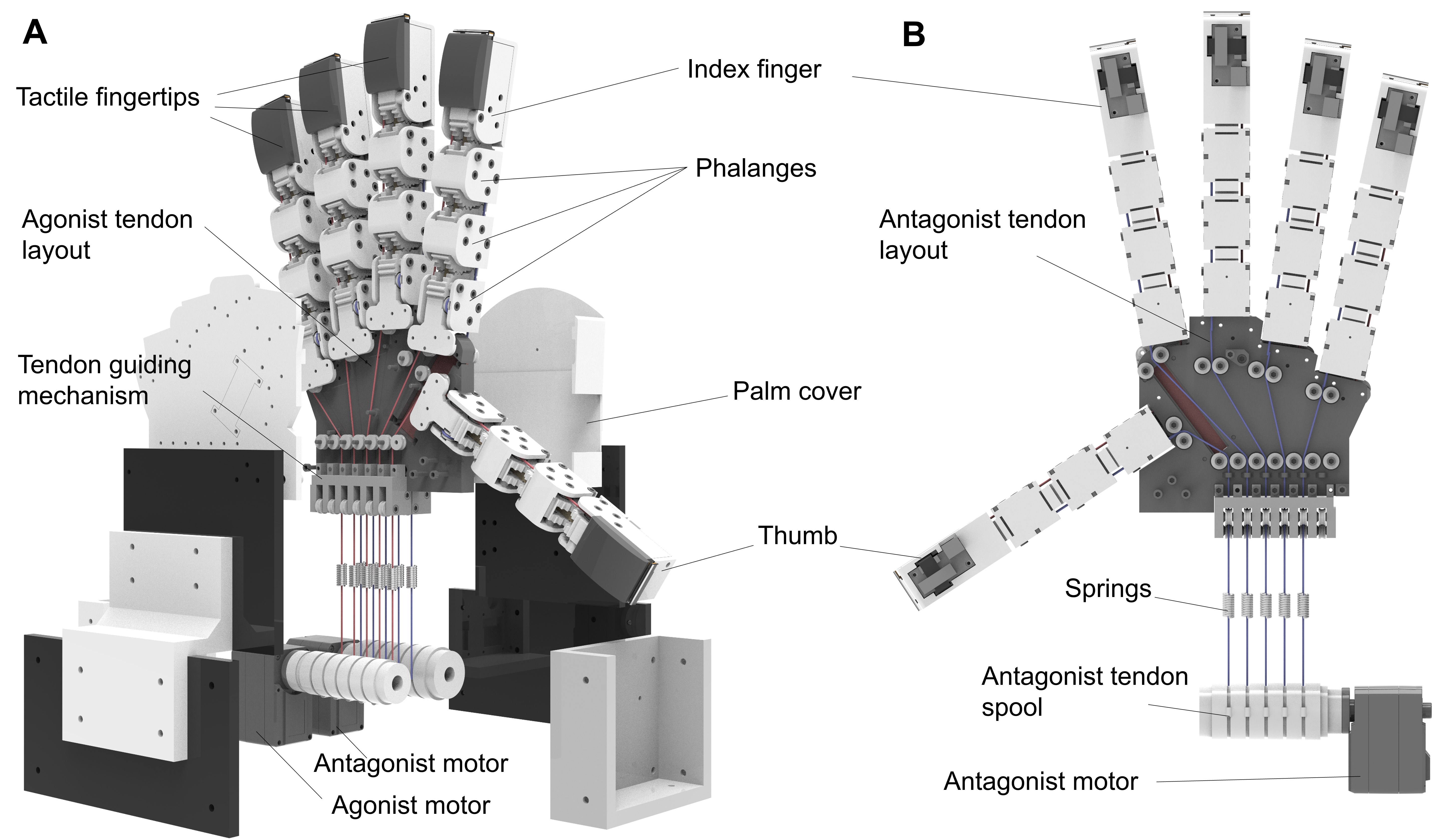} \\
	\end{tabular}
	\caption{Exploded Front view (A) and back view (B) of the Tactile SoftHand-A. The main components shown include the agonist tendons (in red), antagonist tendons (in blue), differential mechanism due to the tendon layouts and couplings, agonist and antagonist motors, phalanges and tactile fingertips. Inside each joint is a connecting rod to maintain correct gear engagement. The overall length of Tactile SoftHand-A is 200\,mm with palm width 90\,mm and length 26\,mm from palm to back cover.}
	\label{fig_hand_exploded}%
	\vspace{0em}
\end{figure*}

The Pisa/IIT SoftHand~(\cite{catalano2014adaptive}) exemplifies these aspects of robot hand design: it is  a highly-adaptive anthropomorphic tendon-driven hand that has naturalistic movements characteristic of human hand motion driven with just one motor.  The hand is highly adaptive due to the soft robotic principle of adaptive synergies~(\cite{bicchi2011modelling}) that synchronizes the motion of the fingers due to a common tendon looping through all the fingers~(\cite{catalano2014adaptive}). As a consequence, the hand exhibits dexterity comparable to far more highly-actuated hands, such as being able to grasp various shaped objects and even basic tool manipulations when the hand is mounted on a robot arm~(\cite{della2017quest}). This dexterity is enhanced with tactile sensing in the fingertips~(\cite{ford2024shear}), enabling grasp-manipulation capabilities such as handling a flexible cup without crushing or dropping it under sudden changes in object weight, and being able to react to disturbances of held objects, {\em e.g.} in reaction to a human guiding the object motion. 

An important direction to progress robotic hand research is to improve the dexterity of underactuated hands while retaining the benefits of simplified designs and eased control. In our view, the Pisa/IIT SoftHand gives a foundation for such progress, being anthropomorphic and exhibiting relatively high dexterity for a single motor. However, its original design is not easy to customize, for example to add actuators or to change the tendon routing; also, it is fabricated by injection molding. Inspired by the Yale GrabLab OpenHand series of 3D-printable customizable grippers~(\cite{ma2017yale,odhner2014compliant}), we designed the BPI (Bristol/Pisa/IIT) SoftHand that is entirely 3D-printed~(\cite{li2022brl}). For easier fabrication, the joints were simplified to pivot around linkage attachments, and soft synergies (springs) in the tendon-actuator coupling recovered some adaptivity and force transmission sacrificed by higher friction in the tendon routing due to the manufacturing.


However, in researching highly-underactuated tendon-driven hands, we noticed that a key but underappreciated design aspect is the extension, or reset, of the finger joints. Clearly, the human hand has different tendons to close and open the hand according to agonist and antagonist muscle groups. Yet almost all tendon-underactuated robot hands use passive elastic elements~(\cite{makino2017high,martin2014design,salvietti2018co}) such as rubber bands~(\cite{catalano2014adaptive,li2022brl,ma2017yale}), torsion springs~(\cite{7225155,8078253,jiang2014modular,mizushima2018multi,stuart2017ocean, aukes2014design, chen2020underactuation}) or soft continuous materials~(\cite{7993072}) to reset the finger joints~(\cite{6663692}). As we will see later in this paper, an ever-present passive restoring force reduces the reactivity, dexterity, and load bearing of the hand. 

This paper investigates the design and development of antagonistic tendon mechanisms within highly underactuated anthropomorphic robot hands. Furthermore, we take the view that the dexterity of the hand due to its actuation cannot be separated from that due to its control, which depends on the integration and deployment of tactile sensing. Therefore, we introduce an entirely 3D-printed, highly-underactuated, five-finger robotic hand named the Tactile SoftHand-A (Figure~\ref{fig_hand_exploded}), which features two actuators: one each to open and close the hand. Within the hand structure, we introduce a new design of fully 3D-printed tactile sensor that requires minimal hand assembly and is printed directly within the robotic fingertip. We show that the combination of an antagonistic tendon mechanism and fingertip tactile sensing significantly improves the hand's dexterity compared to the BPI SoftHand without these features.

An important improvement to the hand's dexterity emerges from this antagonistic mechanism: it then becomes possible to actively control the distal interphalangeal (DIP) or proximal interphalangeal (PIP) joints of the fingers. This control is by coordinating the agonist and antagonist motors, as humans also seem to do when grasping delicate objects with our fingertips. When coupled with the tactile sensing, the antagonistic coordination can then effectively control grasp gestures while maintaining the hand's adaptivity. There is a subtlety, however, that this additional dexterity depends upon the termination of the antagonist tendon and the limited use of passive elements. Another benefit is that active antagonism, by eliminating passive elastic elements, offers lower resistance during the hand-closing process, thereby increasing efficiency, responsiveness, and gripping strength.

Overall, our \textbf{main contributions} in this work are:\\
\noindent 1) We introduce a class of antagonistic tendon mechanisms suitable for the fingers of highly-underactuated robot hands. These finger designs differ in the termination of the antagonist tendons, which determines whether the DIP or PIP joints can be actively controlled, respectively. Together, they constitute examples within a larger design space of finger structures and tendon mechanisms.\\
\noindent 2) We develop a fully 3D-printed tactile fingertip that combines multi-material and multi-hardness components within the mechanical structures of the phalange, eliminating the need for additional fabrication steps. Additionally, we develop a simple tactile model capable of detecting contact and slip. We are the first to introduce this fabrication method because we introduce a novel 3D-printing step that mixes clear support with other materials to adjust their softness.\\
\noindent 3) We combine the above to build a novel, underactuated robot hand with antagonistic tendons, tactile fingertips and new differential mechanisms: the Tactile SoftHand-A. We also develop a control scheme based on analytically extracting tactile features that integrates with the antagonistic tendon mechanism to deploy the functionality of this hand.\\ 
\noindent 4) We evaluate the Tactile SoftHand-A's performance, including adaptivity and grasping ability, through finger and hand-synergy experiments. We further evaluate the tactile performance, demonstrating the improved reactivity to contact disturbances, including slip, and how this enables synchronization with a human hand for use in teleoperation.

This paper is organized as follows. The next section briefly describes and compares several representative underactuated robotic hands, highlighting their main features and differences from the Tactile SoftHand-A; additionally, some works related to vision-based tactile sensors are introduced. The third section details the design of our novel fingers equipped with various antagonistic tendon mechanisms, additionally introducing a physical model of the working principle and a workspace simulation. The fourth section describes the design and development of the tactile fingertip and gives a simple analytic model for detecting contact and slip. The fifth section develops the Tactile SoftHand-A, covering the overall design with D-type finger, differential mechanism, and tendon layout. The sixth section details the control scheme, which is followed by two sections on the experimental design and results, covering the evaluation of a single finger, evaluation of the entire hand, grasping adaptivity, and tactile feedback control. Finally, we discuss these results and make our conclusions.




\begin{figure*}[t!]
	\centering
	\begin{tabular}[b]{c}
		\hspace{0cm}\includegraphics[width=1\textwidth,trim={0 10 0 0},clip]{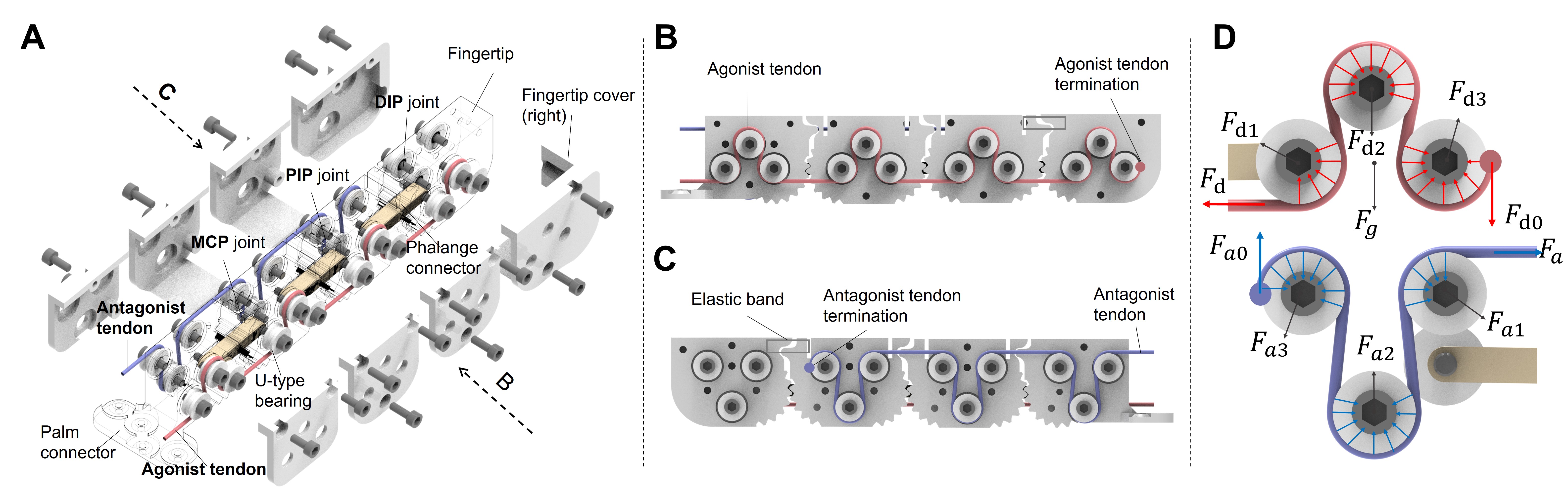} \\
	\end{tabular}
	\caption{The D-type (distal) finger where the antagonist tendon terminates on the DIP joint. (A) Exploded view. (B) Right-side view. (C) Left-side view. (D) Diagram showing the forces for analysis of the motion of the D-type fingertip. Connecting rods within the joints in (A,D) are shown in yellow-brown. The finger's total length, width, and height are 115\,mm, 20\,mm, and 23\,mm, respectively. }
	\label{fig_finger_sketch}%
\end{figure*}

\section{Background and Related work}\label{bgd}

The primary advantage of using passive elastic components lies in the significant reduction of integration and control complexity. However, this approach restricts control accuracy, as the behavior of passive components depends on their pre-set strengths and shapes, which may not suffice for intricate movements. Moreover, unlike motors, passive components cannot adjust their output characteristics, such as force and response speed, via software. In comparison, some robot hands based on a linkage drive excel in finger repeat positioning accuracy and response speed (\cite{nurpeissova2021open}). Nevertheless, they lack adaptability to various object shapes, a challenge that underactuated tendon-driven robotic hands can address~(\cite{catalano2014adaptive}).

\subsection{Tendon-driven Robot Hands}
At present, research on robotic hands has two main directions. One is to try to obtain the full dexterity of the human hand; {\em i.e.} to achieve all human hand functions by near-fully actuating the hand~(\cite{7993072,jacobsen1986design,butterfass2001dlr,liow2019olympic,wang2019eagle,stuart2017ocean}). The other direction is to reduce the complexity and control difficulty of robotic hands through tendon-driven~(\cite{chen2018topology,li2022brl,catalano2014adaptive,mizushima2018multi}), pneumatic-driven~(\cite{9854144}) and linkage mechanisms~(\cite{9137654}). This second approach benefits from progress in soft robotics, where there is morphological intelligence instantiated in the design and coupling of the structure comprising the hand. 

Some near fully-actuated robotic hands seek to achieve complete dexterity, including the Shadow anthropomorphic hand, M2 gripper~(\cite{ma2016m}), the UTAH-HIT hand~(\cite{jacobsen1986design}) and the ROBISS hand~(\cite{7225155}). There are a variety of mechanisms that each have benefits and limitations. For example, the M2 Gripper utilizes a dual-tendon system to switch between underactuated and fully actuated modes, offering flexibility in grasping. However, it relies on passive reset mechanisms such as elastic bands or springs, which limit precise control of finger repositioning after releasing an object. Another approach is to adding electrostatic actuators to the joints of a tendon-driven robotic hand, to increase gripping force and dexterous ability by unlocking specific joints to decouple the finger joints~(\cite{aukes2014design}). However, an agonist force is still needed to overcome the residual braking force and make the joint move. The challenge with all near fully-actuated hands is the complexity of the control needed to deploy the hand to grasp or manipulate objects without damaging it. 

The Pisa/IIT SoftHand~(\cite{catalano2014adaptive}) is inspired by the form of the human hand, but seeks to simplify the control by using just one tendon through all joints that is driven by one actuator. This coupling between the elements of the hand gives it the ability to adapt to the shape of various objects. In addition, it uses a novel differential mechanism as an ingenious way to reduce the number of actuators while still having naturalistic hand movements that are characteristic of human hand motion~(\cite{sun2021design,xiong2016design,bicchi2011modelling,santello2016hand}). These differential mechanisms usually use the parallel slider mechanism~(\cite{sun2021design}), moving-pulley mechanism~(\cite{4543295,gao2021anthropomorphic,chen2014mechanical}), fixed pulley mechanism~(\cite{chen2020underactuation}) or spring groups~(\cite{li2022brl}), which can be used to compose the synergy scheme~(\cite{catalano2014adaptive,6663692,bicchi2011modelling,fan2018research}).

In the process of studying the above robot hands, we found that almost all of the underactuated tendon-driven hands use passive elements such as torsion springs~(\cite{7225155,8078253,jiang2014modular,mizushima2018multi,ma2016m, aukes2014design}), high-strength springs~(\cite{makino2017high,martin2014design,stuart2017ocean}) and elastic bands~(\cite{catalano2014adaptive,li2022brl,ma2017yale,stuart2017ocean}) to achieve joint reset motion. Few underactuated tendon-driven hands use active extension to reset the fingers. Of these, one has proposed a robotic finger with a release tendon and a grasp tendon~(\cite{511778}). However, its bearing and tendon layout limits the rotation angle of its finger joint, since the tendon can detach from the bearing or axis during rotation, so its motion space is relatively small~(\cite{511778,320882}).  Here, we use a different bearing layout to guide the flexion tendon and extension tendon to increase the joint rotation angle of the joint. 

Another example of a robot hand employing a two-tendon mechanism for the robotic finger to enable active flexion and extension is by
~\cite{ruotolo2021grasping}. Additionally, the flexible connections between joints, coupled with the absence of mechanical limits, allow the fingers to bend in reverse, thereby increasing its operational workspace. However, a limitation remains: the joints are still coupled, and active control of the fingertip’s rotation cannot be achieved. Another study proposed a robotic finger based on the biomimetics of human finger muscles, with two actuators for the adductive motion of the joints and control of the extension of the finger through a third actuator~(\cite{shirafuji2014development}). However, that finger has a complex tendon layout that is not conducive to the design and assembly of the overall hand. The fingers of the CLASH hand (\cite{friedl2020clash}) utilize two independent tendon systems to drive the MCP and DIP joints (without a PIP joint), thereby expanding the finger's workspace. Additionally, a differential mechanism couples the DIP joints of two fingers, providing these joints with a degree of adaptability. However, while this design introduces adaptability between the DIP joints, it sacrifices the intrinsic adaptability between the MCP and DIP joints of the individual fingers.

It is also worth noting that in examples of more fully-actuated robot hands, both the DLR Hand II~(\cite{butterfass2001dlr}) and the Shadow Hand utilize modular fingers where the DIP joints are not independently controllable. To reduce the number of required motors, the DIP joints in these hands are coupled with the PIP joints; as a PIP joint flexes, the DIP joint follows proportionally. In contrast, the (fully-actuated) Allegro Hand achieves DIP joint control by incorporating an additional joint motor at the DIP joint. The BPI SoftHand-A, however, actively controls the DIP joint without extra motors, utilizing coordinated control between driving and antagonistic motors. This design enhances the hand's dexterity and grasping capabilities.

Overall, we have observed that currently, no tendon-driven robotic hand can decouple the DIP or PIP joint while ensuring adaptive capability, nor actively control the DIP or PIP joints to adjust the grasping gesture while maintaining adaptive grasping ability. 

Therefore, here we propose a novel tendon-driven finger design that achieves active extension and control of the DIP or PIP joints through coordinated control of agonist and antagonist tendons. Based on this new design, we have also developed a novel robotic hand that with just two motors can actively manage closing, opening, and grasping gestures.

\subsection{Tactile Sensing in Robot Hands}

Alongside the design and fabrication of the anthropomorphic hand, is its capability to be controlled, either autonomously or through human teleoperation. In this respect, having tactile sensing capabilities in the fingertips, and possibly over the rest of the hand, is necessary to provide direct information about contact. Historically, there have been many ways of integrating tactile fingertips into robot hands (\cite{kappassov2015tactile}). 

Compared to traditional electronic-based tactile sensors, vision-based tactile sensors (VBTS) using cameras have become prominent for various reasons, including the recent miniaturisation and cost reduction in camera technology and the high-resolution nature of the sensory data fitting well with advances in computer vision. The tactile images can provide rich physical information for robotic interaction tasks, such as texture features and object pose. Researchers have developed many VBTS, such as MIT's GelSight (\cite{yuan2017gelsight}), and BRL's TacTip (\cite{ward2018tactip}, \cite{lepora2021soft}) and BioTacTip (\cite{li2024biotactip}). 

However, the manufacturing processes for these vision-based tactile sensors are complex and labor-intensive, typically exceeding many hours (\cite{zhang2022hardware}). Many such sensors derived from the GelSight, such as the 9DTact~(\cite{lin20239dtact}), require specific molds and the casting of silicone to create soft skins. Also, the use of transparent windows in all VBTSs mentioned above requires laser equipment for cutting and hand-placement/gluing the window. In manufacturing the TacTip, transparent gel is injected into the skin and window cavities, and specialized vacuum equipment is necessary to remove any bubbles that may form during injection and curing~(\cite{ward2018tactip}). Furthermore, when integration within robot hands is needed, researchers must create new molds and produce connection parts to the phalanges or palm~(\cite{zhang2024palmtac,zhang2024tacpalm}). These manual manufacturing processes significantly restrict the form-factor of vision-based tactile sensors, as well as the variety and distribution of inner structures such as markers. Furthermore, the manufacturing variation inherent in manual production can impact the generalization performance of perception systems using the tactile images to predict contact features, as the perception models can vary even across fingertips on the same robot hand (\cite{ford2024shear}).

Therefore, we have developed a new printing technology that allows for the 3D printing of multi-material and multi-component parts without any post-fabrication steps such as gel injection or glue windows. Using this technology, we have for the first time integrated VBTS with the phalanges of a robotic hand in a single print.

\begin{figure*}[t!]
		\centering
		\begin{tabular}[b]{c}
			\hspace{0cm}\includegraphics[width=1\textwidth,trim={0 0 0 0},clip]{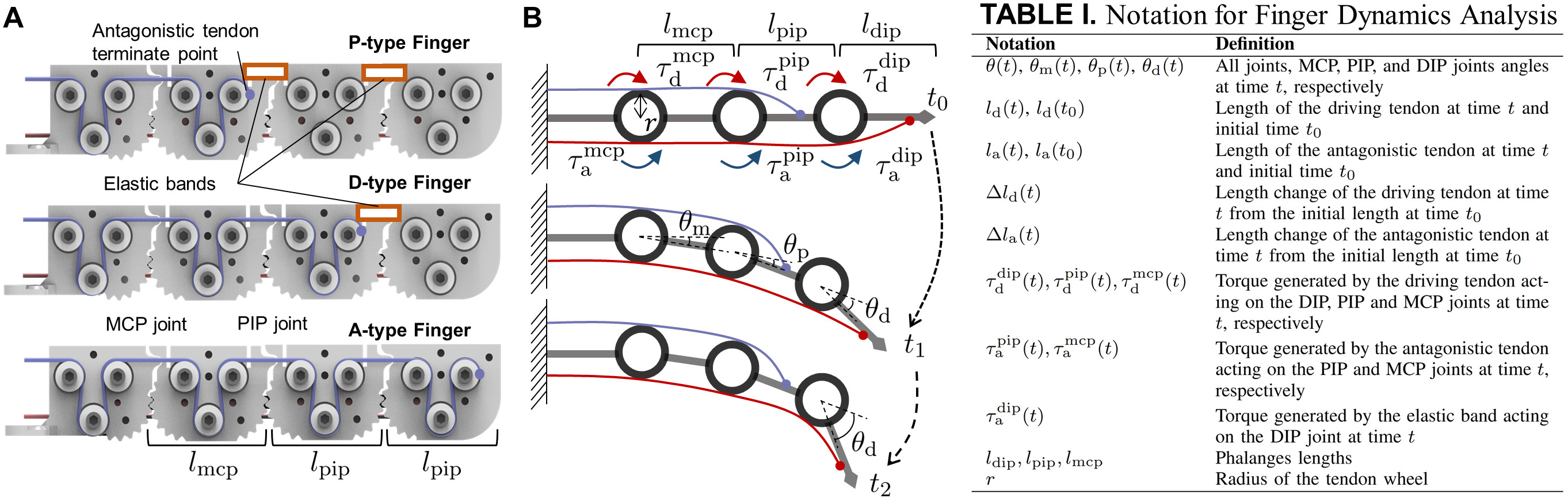} \\
		\end{tabular}
             \caption{Antagonist tendon comparison for P-type (proximal-type), D-type (distal-type) and A-type (antagonstic-type) fingers. (A)~Side views. (B) Kinematically-equivalent model of D-type finger showing the main variables of interest (described in inset Table).}
		\label{fig_fingers_compare}%
		\vspace{1em}
	\centering
	\begin{tabular}[b]{c}
		\hspace{0cm}\includegraphics[width=1\textwidth,trim={0 0 0 0},clip]{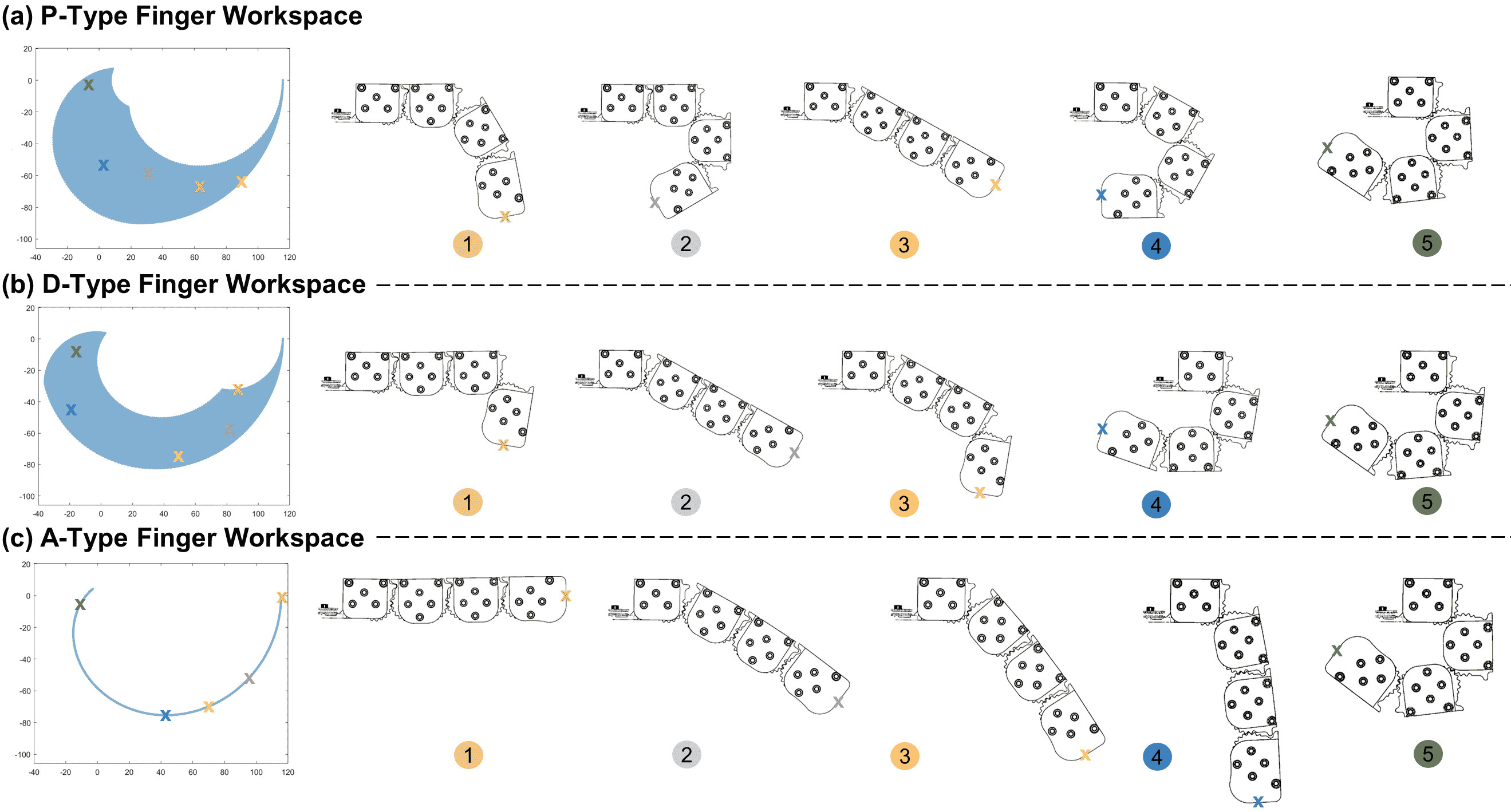} \\
	\end{tabular}
	\caption{Workspace comparison of the P-type, D-type and A-type fingers. The left shaded regions show the entire workspaces for the tip of the fingers, with illustrative finger shapes within those workspaces shown to the right. Note the larger workspaces of the P-type and D-type fingers due to the capability to control the PIP and DIP joints, compared to that of the A-type finger.}
	\label{fingers_w_compare}%
\end{figure*}

\section{Development of Robot Fingers Featuring Active Antagonism}\label{fingers}
Based on different tendon configurations, we have developed three types of fingers, each exhibiting distinct flexion gestures, termed the A-type finger, D-type finger and P-type finger for fully antagonistic-type (A), distal-type (D) and proximal-type (P), respectively (see Figures~\ref{fig_fingers_compare}(A), \ref{fingers_w_compare} and \ref{fig_finger_gestures}). The A-type finger, which features an antagonist tendon running through the entire finger, exhibits the fastest extension. The D-type finger features active control of the distal interphalangeal (DIP) joint, achieved through the coordination of agonist and antagonist tendons and a passive element on the DIP joint. Similarly, the P-type finger employs a comparable mechanism to actively control the proximal interphalangeal (PIP) joint. Their design details are as follows.

\subsection{Design of A-type (antagonistic-type) finger}\label{atype}

As shown in Figure~\ref{fig_finger_sketch}(D), the bearings used for winding the agonist tendon are arranged in an upright triangular configuration, while the bearings for winding the antagonist tendon are arranged in an inverted triangular configuration. This arrangement is designed to generate a greater downward force on the tendons in the middle pulley, thereby applying sufficient torque to the joints to drive joint rotation or reset the joint. In this study, the pulley arrangement is the same for all types of fingers, with the height of the middle pulleys on each phalanx kept consistent. Due to friction and tendon deformation, the MCP joint typically experiences the highest torque first, resulting in its rotation, which provides the finger with a larger workspace and greater adaptability (this phenomenon is demonstrated in the multimedia video file). To achieve specific finger gestures ({\em e.g.} the synchronous rotation of all joints in \cite{megaro2017designing}), it is necessary to adjust the height of the middle pulleys on each phalanx. The arrangement of the pulley system for the antagonist tendons follows the same principle and similar considerations apply.

Similarly to the design of the BPI-SoftHand~(\cite{li2022brl}), a connecting rod at each joint links the adjacent phalanges to maintain correct gear engagement and prevent dislocation (differing from the original Pisa/IIT-SoftHand design of~\cite{catalano2014adaptive}).

In contrast to the D-type and P-type fingers, the antagonist tendon of the A-type finger terminates at the fingertip (Figure~\ref{fig_fingers_compare}(A)). This means that the A-type finger does not require an elastic band or other passive element to assist its extension, resulting in a simpler structure and full active antagonism. Consequently, during finger flexion, the absence of resistance from an elastic band gives a greater fingertip force and faster reaction compared to the other two types of fingers. However, despite these advantages, the A-type finger lacks active control capabilities for the DIP or PIP joints, unlike the other two types, the effect of which is assessed experimentally below.

\subsection{Design of D-type (distal-type) finger}\label{dtype}
As illustrated in Figure~\ref{fig_finger_sketch}(A), the D-type finger primarily consists of phalanges, phalangeal coverings, tendons, and a bearing mechanism. The phalanges are composed of a fingertip, two middle phalanges, and a base phalanx, interconnected through gear engagement to form three rotary joints: the Distal Interphalangeal (DIP), Proximal Interphalangeal (PIP), and Metacarpophalangeal (MCP) joints. A bearing mechanism with multiple U-type grooves is also affixed to the four phalanges. The agonist tendon (Figure~\ref{fig_finger_sketch}(A), red cable) and antagonist tendon (Figure~\ref{fig_finger_sketch}(A), blue cable) run through the entire finger via the U-type grooves on the bearings. For instance, bearings arranged in an equilateral triangle on the right side of the middle phalanx facilitate flexion movement by providing agonist force to the agonist tendon. Conversely, bearings in an inverted triangle arrangement on the left side support extension movement. Notably, the antagonist tendon of the D-type finger terminates at the first middle phalanx. Therefore, the force required for resetting the DIP joint is provided by an elastic band positioned between the fingertip and the middle phalanx, coupled to the active reset of the other finger joints.

\subsection{Design of P-type (proximal-type) finger}\label{ptype}
The primary structure of the P-type finger is identical to that of the D-type finger. The only difference lies in the configuration of the antagonist tendon (see Figure~\ref{fig_fingers_compare}). Unlike the D-type finger, the antagonist tendon of the P-type finger terminates at the second middle phalanx. Consequently, it is necessary to incorporate passive components at the DIP and PIP joints to facilitate the repositioning of the corresponding phalanges, where we use elastic bands similar to the D-type finger. It is worth noting that the position of the elastic bands significantly affects the amount of resistance they generate at the joints, which influences the gesture of the finger. For the D-type finger, the DIP joint is fully decoupled and thus remains unaffected by the elastic bands. However, for the P-type finger, the PIP and DIP joints are still connected via elastic bands, meaning that the relative rotation between these joints is influenced by the stiffness of the elastic bands. The farther the elastic bands are positioned from the joint's rotational center, the greater their deformation during rotation, which in turn increases the resistance applied to the joint. Therefore, by adjusting the position of the elastic bands on DIP or PIP joint independently, the joint resistance can be set for precise regulation of the finger's flexion gesture.

\subsection{Kinematic Analysis of D-type Finger} \label{force}

The D-type finger design features active control of the DIP joints of the fingers at any position through coordinated control of the agonist and antagonist tendons. To examine this feature, we adopt an analytic model in which the force distribution and the schematic of the finger are shown in Figures~\ref{fig_finger_sketch}(D) and \ref{fig_fingers_compare}(B) respectively, with symbol notation introduced in Table I. We consider two time periods for the movement of the D-type finger: the first without active control of the DIP joint from time $t_0$ to $t_1$, and the second with active control of the DIP joint from $t_1$ to $t_2$.

The finger rotation angle $\theta(t)$ is given by the sum of DIP rotation angle $\theta_{\rm d}(t)$, PIP rotation angle $\theta_{\rm p}(t)$, and MCP rotation angle $\theta_{\rm m}(t)$:
\begin{equation}
\theta(t) = \theta_{\rm m}(t) + \theta_{\rm p}(t) + \theta_{\rm d}(t).
\label{eq:finger_rotation}
\end{equation}
Accordingly, the length change of the agonist and antagonist tendons is represented by $\Delta{l_{\rm d}}(t)$ and $\Delta{l_{\rm a}}(t)$, where
\begin{equation}
\Delta{l_{\rm d}}(t) = \left|{l_{\rm d}}(t) - {l_{\rm d}}(t_0) \right|,
\label{eq: tendon1}
\end{equation}
\begin{equation} 
\Delta{l_{\rm a}}(t) = \left|{l_{\rm a}}(t) - {l_{\rm a}}(t_0) \right|.
\label{eq: tendon2}
\end{equation}

First, let us consider a period from $t_0$ to $t_1$, in which the antagonism synchronizes with the agonist tendon. Therefore, there is no resistance acting on the MCP and PIP joints: 
\begin{equation} 
\tau_{\rm a}^{\rm mcp}(t_1) = \tau_{\rm a}^{\rm pip}(t_1) = 0,
\label{eq: t1_torque}
\end{equation}
\begin{equation}
\begin{aligned}
\Delta{l_{\rm d}}(t_1) - \Delta{l_{\rm a}}(t_1) =  \theta_{\rm d}(t_1) r.
\end{aligned}
\label{eq: t1}
\end{equation}
Now, in a second period from $t_1$ to $t_2$, we use active antagonism to control the DIP joint. We keep the antagonist tendon at the same length and pull the agonist tendon. This means that the torque acting on the MCP and PIP joints by the agonist tendon is equal to the torque acting on the MCP and PIP joints by the antagonist tendon, and is less than the maximum stall torque of the antagonist motor, $\tau_{\rm a}^{\rm stall}$. Thus, the MCP and PIP joints will remain in their current state:
\begin{equation}
    \tau_{\rm d}^{\rm mcp}(t_1) + \tau_{\rm d}^{\rm pip}(t_1) = \tau_{\rm a}^{\rm mcp}(t_1) + \tau_{\rm a}^{\rm pip}(t_1) \leq \tau_{\rm a}^{\rm stall},
\end{equation}
\begin{equation}
\theta_{\rm m}(t_1)\!=\!\theta_{\rm m}(t_2),\, \theta_{\rm p}(t_1)\!=\!\theta_{\rm p}(t_2),\, \Delta{l_{\rm a}}(t_2)\!=\!\Delta{l_{\rm a}}(t_1).
\end{equation}
Hence, the difference in length of the agonist tendon between $t_1$ and $t_2$ is
\begin{equation}
\begin{aligned}
\Delta{l_{\rm d}}(t_2) - \Delta{l_{\rm d}}(t_1) &= (\theta_{\rm d}(t_2)-\theta_{\rm d}(t_1))r,
\end{aligned}
\label{eq: t2}
\end{equation}
which results in bending of the DIP joint.

Therefore, we can actively control the DIP joint by controlling the displacement of the agonist and antagonist tendons.

\begin{figure*}[htbp]
	\centering
	\begin{tabular}[b]{c}
	\hspace{0cm}\includegraphics[width=1\textwidth,trim={0 20 0 0},clip]{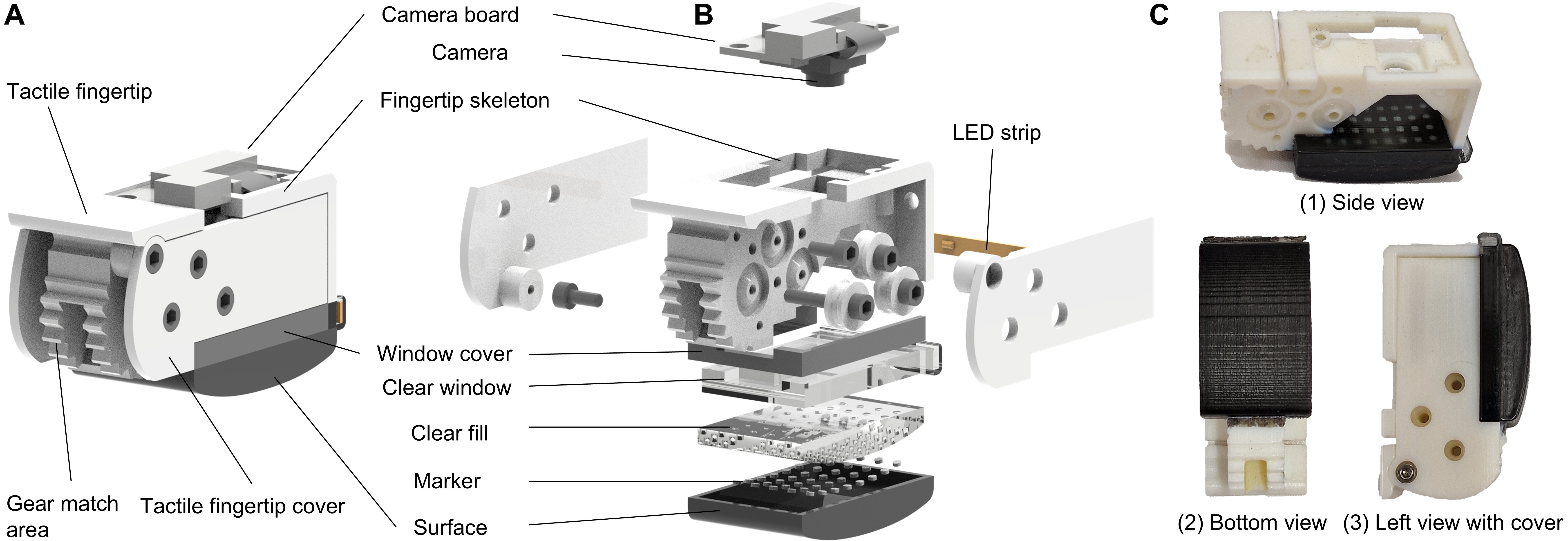} \\
	\end{tabular}
	\caption{Construction and design of the fully 3D-printed tactile fingertip. (A) side view. (B) Exploded view. (C) Print and construction of the tactile fingertip. The only hand fabrication is to insert the camera board in the top on the fingertip. }
	\label{fig_tac_fingtip}%
	\vspace{1em}
	\centering
	\begin{tabular}[b]{c}
		\hspace{0cm}\includegraphics[width=1\textwidth,trim={0 10 0 0},clip]{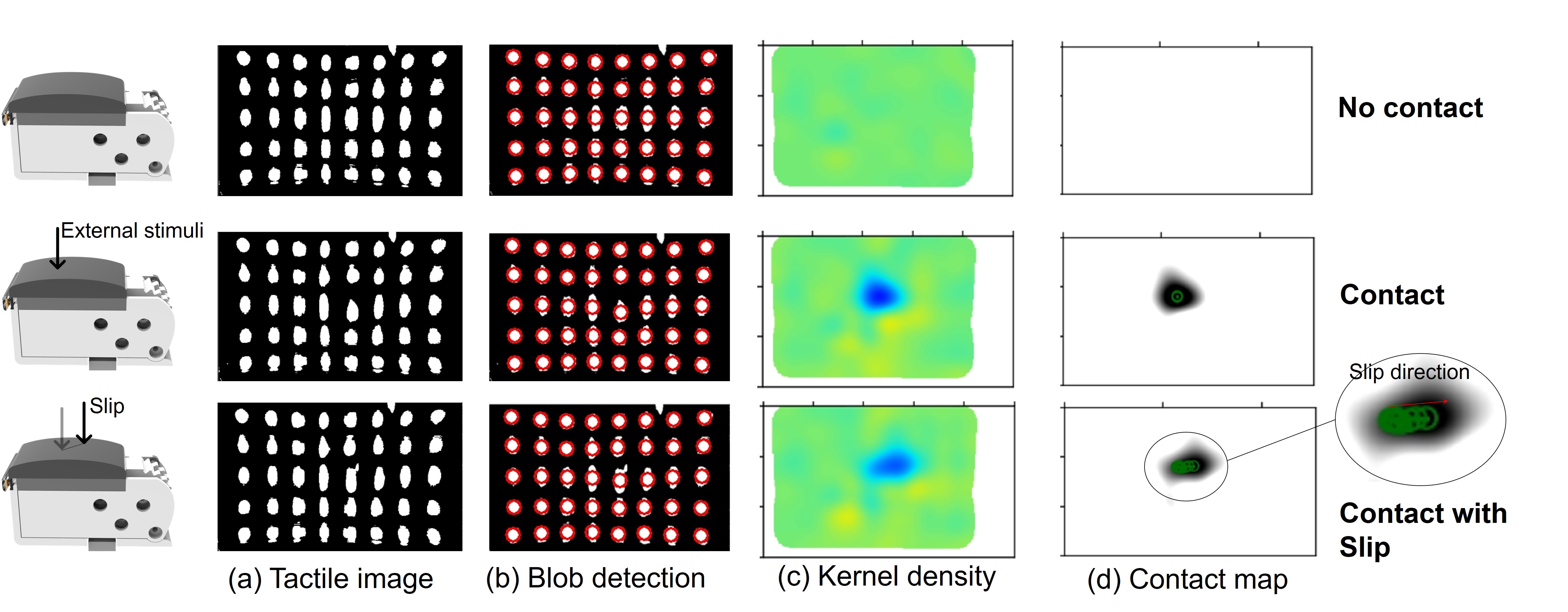} \\
	\end{tabular}
	\caption{Tactile image processing under contact and slip for three consecutive times $t_i$. (a) Raw grayscale tactile images. (b)~Marker detection using the DoH blob-detection method (red circles). (c) Contact map by applying a Gaussian kernel density transformation on the marker centroids, with the contact region visible in blue. (d) Extracted contact centroid (middle row, green circle) and effect of contact slip (bottom row, green circles and red arrow). }
	\label{fig_images_process}%
	\vspace{0em}
\end{figure*}

\subsection{Simulation of Finger Workspace}\label{compa}

After demonstrating that the A-type, D-type, and P-type fingers can execute diverse gestures through the coordinated control of the agonist and antagonist tendons, an important comparison is their workspaces. To this end, we utilized the MATLAB Robotics Toolbox to construct models for these finger types, employing the Denavit-Hartenberg (DH) matrix. This serves as a preliminary step to validate the overall design concept and predict the workspace of different finger-type configurations. This enables us to systematically compare the workspaces of different finger types under the same conditions, which would otherwise be highly time-consuming and challenging to implement using physical prototypes for each design iteration.

For example, the D-type finger requires an independent control input for active management of its DIP joint due to its capability for autonomous movement; meanwhile, the proximal PIP and MCP joints are interconnected, necessitating a separate control input to adjust the angles of both PIP and MCP joints. By applying the Monte Carlo method to generate varied inputs, we employ forward kinematics to delineate the fingertip's workspace. 

Using the above method, we plotted the fingertip vertex positions for various gestures in the simulator (Figure~\ref{fingers_w_compare}). Unlike the D-type and P-type fingers, the A-type fingers lack active control over the DIP or PIP joints, resulting in a workspace confined to a fixed arc due to the coordinated action of the agonist and antagonist tendons (Figure~\ref{fingers_w_compare}(c)). Notably, the D-type finger demonstrates a more uniform workspace distribution, particularly in extended gestures, leading to the selection of D-type configured fingers for testing and developing the Tactile SoftHand-A, as discussed in the later sections of this paper. 

It is worth noting that the above simulation was completed without touching any objects or obstacles. Because the fingers have adaptive capabilities, simulating workspace changes in contact with objects or obstacles is not discussed in this section, but is evaluated experimentally later instead.

\section{Development of Integrated Tactile Fingertips}

The fingers are intended to have vision-based tactile sensing in the distal phalanges. Here we introduce a novel design where the entire sensing structure of the fingertip is 3D-printed as a single component integrated into the physical structure of the finger, requiring no manual fabrication except for the insertion of the camera module. First, we detail the main components and design considerations of the fully-integrated fingertips, then their fabrication utilizing advances in multi-material 3D-printing technology.

\subsection{Overall Design}
As depicted in Figure~\ref{fig_tac_fingtip}, the tactile fingertip comprises three modules: the perception module, the contact module, and the actuation module. The perception module is composed of a camera and an LED strip: the camera captures high spatial resolution tactile images, while the LED strip mounted on the front of the tactile fingertip skeleton provides consistent lighting inside the contact module. The actuation module, consisting of bearings and a tendon mechanism installed on either side of the tactile fingertip skeleton, provides torque for fingertip rotation. The contact module components include the fingertip skeleton, a clear window, window cover, soft transparent filling material, markers, and the skin surface. The fingertip skeleton integrates a gear match area to form a gear pair with the second phalanx. 

The perception and actuation modules are also mounted on the phalangeal skeleton, with bearings and the tendon mechanism secured by bolts on both sides. The camera lens and camera board are affixed to the top of the fingertip. The skin surface, located at the fingertip's pad, converts external contacts into internal deformation, thereby moving markers attached on pins to its inner surface based on the TacTip design~(\cite{ward2018tactip,lepora2021soft}). These are then captured by the camera to form tactile images. Additionally, the black outer skin surface and window cover isolate external light interference, to aid in producing high-quality tactile images. The soft, transparent clear fill is designed to increase the surface's deformation range while providing internal support and resetting its deformation. The rigid, transparent clear window serves to support the clear fill and refract light from the LED strip.

\subsection{Fabrication of the Tactile Fingertip}
The perception and actuation module components of the tactile fingertip can be purchased inexpensively, and the remaining components produced using multimaterial 3D-printing (Stratasys J826 Prime 3D-printer). The entire assembly of the contact module's components, including the flexible black opaque surface and pins (soft Agilus30 material), hard white markers (hard VeroWhite material), soft transparent filling (soft Agilus30 clear mixed with support material), hard transparent window (hard VeroClear material), hard black opaque window cover, and the hard white skeleton (hard VeroWhite material), is integrally formed as a single component without any other post-fabrication processes required. 

The new capability to 3D-print the assembly of the contact module as one piece offers significant benefits in ease of assembly and fabrication repeatability. In previous versions of the TacTip~(\cite{lepora2021soft}), after printing we needed to clean the support, glue the acrylic window, inject gel (requiring mixing then degassing in a vacuum chamber), then leave in a dryer for 10 hours or more before it could be used. By using this new technology, we can quickly produce tactile fingertips, from design to application in just 1 hour. 

This capability is made possible by an innovation we have introduced in mixing support material with Agilus material to print a clear fill that, while maintaining its transparency, can have its hardness adjusted by altering the proportion of support material. Previously, the hardness of the pure Agilus material available with the printer was found to be excessively high for use as a soft filler. Usually, the support material is required for the layering of complex structures and is cleaned post-printing. We noticed that one of the support materials for complex pieces happens to be optically clear, and through experimenting, found it is compatible with mixing with other materials. Hence, we use the support material to adjust the filling material's softness to a level that would otherwise not be possible using commercial 3D-printing materials.

Other possibilities to extend these techniques would be to use solid-liquid hybrid 3D printing, such as proposed by \cite{maccurdy2016printable}. However, those methods use liquid fillers and require additional support materials to prevent interaction between uncured liquid and the deposited photopolymer, which is somewhat more complicated than our approach. With advances in 3D-printing technology, we expect better techniques to become possible for fabricating diverse optical tactile sensors.

\subsection{Tactile Localization Model}
The camera images captured with the perception module are then processed with a tactile model (Figure~\ref{fig_images_process}). Here we are primarily concerned with predicting and using the contact region and its center point, which we will integrate into the robot hand control. The model details are described below.

\subsubsection{Image Preprocessing:}
Initially, the image captured by the camera is cropped to the region in which markers are visible. Subsequently, binary processing is applied to the cropped image, using thresholding on the pixel intensity (set to 180 on a 0 to 255 scale). These preprocessing steps yield a binarized image $I_{\rm gray}$ in which the markers are clearly visible.

\subsubsection{Marker Detection:}
We employ the Determinant of Hessian (DoH) estimation method for marker detection to identify white markers in images. The DoH estimation method is a widely-used technique in image processing to identify blob features. For a given image $I_{\rm gray}$, it operates by maximizing the determinant of the Hessian matrix over image points $(x,y)$, using the matrix of second partial derivatives in the $x$ and/or $y$ directions. To reduce noise and aid feature detection across scales, Gaussian kernel smoothing is applied to the image before computing the matrix. Marker detection is achieved by identifying the locations in the image where the Hessian matrix attains local maxima, denoted by $(x_m,y_m)$ with $m$ the marker index. 

\subsubsection{Contact Region Estimation:}
Markers within the contact region are influenced by external contacts, levering them towards adjacent uncontacted regions, thereby diminishing the marker density in the region of contact. Consequently, the location of the contact region can be inferred by analyzing the local marker density. Here we adopt a methodology involving the following steps: upon identifying the markers $(x_m,y_m)$, we compute the marker density using a Gaussian kernel density estimation method (\cite{silverman2018density, lloyd2024pose, lu2024dexitac}). This approach positions Gaussian kernels at the marker centroids, with a constant kernel width $h$ taken from the mean distance between neighboring markers:
\begin{equation}
\bar{d}(x, y) = \frac{1}{M} \sum_{m=1}^{M} \frac{1}{\sqrt{2\pi}h^2} \exp \left( - \frac{\lVert (x, y) - (x_m, y_m) \rVert^2}{2h^2} \right),
\end{equation}
where the square-distance norm is $\lVert (x, y) - (x_m, y_m) \rVert^2 = (x - x_m)^2 + (y - y_m)^2$. We can see from Figure~\ref{fig_images_process}(c) that the extent of the low marker density areas in the kernel density map (blue region) depends on the contact depth and represents the contact area. Finally, we calculate the lowest density point within that region:
\begin{equation}
(x_{\rm c}, y_{\rm c}) = \underset{(x,y) \in \mathbb{R}}{\mathrm{arg\,min}}\ \bar{d}(x, y),
\end{equation}
which, as shown in Figure~\ref{fig_images_process}(d), represents the contact center (shown as a green circle).

\begin{figure*}[t!]
	\centering
	\begin{tabular}[b]{c}
		\hspace{0cm}\includegraphics[width=1\textwidth,trim={0 0 0 0},clip]{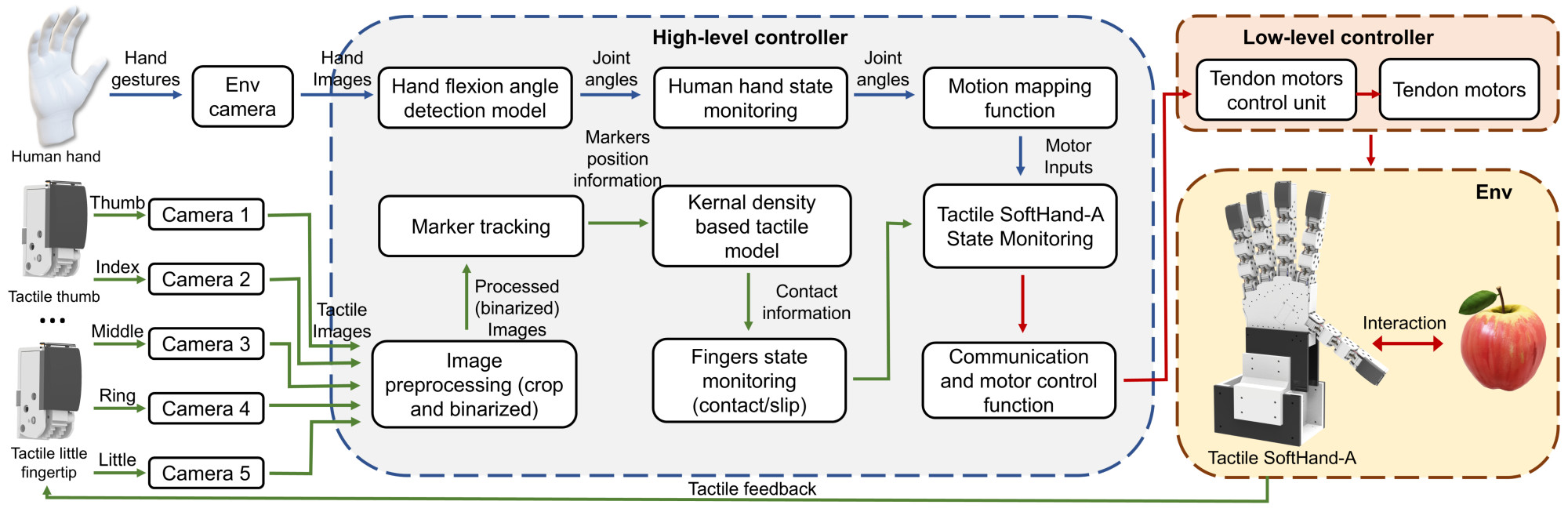} \\
	\end{tabular}
	\caption{Schematic diagram for the processing modules and tactile feedback grasping control flow chart, which are used to control the hand both in open loop from human gestures and closed loop from the tactile feedback.}
	\label{fig_hand_control}%
	\vspace{0em}
\end{figure*}

\section{Development of the Tactile SoftHand-A}\label{hand_Development}

\subsection{Overall Design}\label{overall design}
The Tactile SoftHand-A is a major advance over the previous design of the BPI-SoftHand~(\cite{li2022brl}), with its use of novel D-type fingers featuring active antagonism, coupled with a modified two-layer palm design for distinct routing of agonist and antagonist tendons, and utilization of two differential mechanisms (Figure~\ref{fig_hand_exploded}). Earlier in this paper, we introduced three different types of novel fingers; here, we select the D-type fingers for their more evenly distributed workspace to use in constructing the Tactile SoftHand-A. Overall, the hand has 15 degrees of freedom and 2 degrees of actuation, with each finger encompassing three joints: MCP, PIP, and DIP, marking an evolution in design and functionality from the previous BPI SoftHand.  

The parts of each finger are depicted in the exploded design illustrations in Figure~\ref{fig_finger_sketch}. The hand employs two distinct tendon schemes for the operation of the fingers: one for flexing and another for extending. This dual-tendon system features a soft synergy mechanism, linking each of the five tendons via a spring to its coupling mechanism (tendon spool) and to the differential mechanism, which then connects to a shared spool on the motor, ensuring coordinated and smooth motion.

\subsection{Differential Mechanism}
The Tactile SoftHand-A has two differential mechanisms: one for the active flexing movement of the fingers and one for their active extension movement. Each differential mechanism consists of a soft synergy scheme that couples the tendons for flexion and extension separately to springs, with each mechanism connected to a separate tendon coupling driven by a servo motor (see Figure~\ref{fig_hand_exploded}). Through the combined differential mechanism, the entire hand can be actively opened and closed with two motors. Hence, the synchronization of all fingers can be achieved easily, and grasping adaptability is also kept, as will be verified in the experimental results. 

\subsection{Palm Tendon Layout}
To mitigate friction during tendon force transmission, bearing groups were strategically positioned within the palm, as shown in Figure~\ref{fig_hand_exploded}(B). The setup for the index finger's antagonist tendon illustrates this approach: a bearing set guides the tendon at the finger's base, preventing direct contact with the base phalanx cover. This arrangement ensures the tendon follows a defined path to the bottom of the palm, where it connects to the guiding mechanism and subsequently to a spring, thus avoiding palm contact and minimizing friction during operation.

\section{Human-guided Control Scheme} \label{control scheme}
We developed a comprehensive control strategy for the Tactile SoftHand-A, employing both open-loop and closed-loop control mechanisms. In the closed-loop strategy, the system converts human gestures into motor control signals, utilizing both high-level and low-level controllers (a PC and motor control board). The high-level controller interprets real-time images from the human hand and tactile fingertips, calculating the necessary joint angles, while the low-level controller translates these signals into motor actuation. This dual-level control system allows dynamic adjustments with tactile feedback, ensuring smooth transitions between human gesture synchronization and object anti-slip grasping.

When the robotic hand is not in contact with an object, mimicking human hand gestures simplifies control complexity, as demonstrated in \cite{gioioso2013mapping}. Upon contact, the system switches to closed-loop control, using tactile feedback to automatically adjust the grasping force according to sensor input, preventing slippage. The tactile feedback system, central to the closed-loop control, adapts the hand’s grasping force in response to changes in tactile sensor input. By tracking slippage through marker displacement in tactile images, the system can autonomously adjust the grasp to ensure a secure hold without overexerting force. These adaptive adjustments prevent damage to delicate objects while maintaining control stability.

The control system is an integrated set of modular PID controllers. Specifically, the system consists of four main components (Figure~\ref{fig_hand_control}): a human-hand gesture detection module, a high-level controller (PC), a low-level controller (motor control board), and the Tactile SoftHand-A hardware. The PC processes raw images of the human hand captured by an environment camera, along with tactile images from the fingertips of the Tactile SoftHand-A. These images, combined with gesture and tactile models, generate motor control parameters transmitted to the motor controller. The motor controller then uses PID control to adjust the rotation of two servo motors, enabling the Tactile SoftHand-A to interact with its environment and adjust its grasp based on real-time tactile feedback.

\subsection{Open-loop Control Strategy}\label{exp_ab_control}
As will be described in more detail in the experiments, we will adopt open-loop feedback control to compare the P-type, D-type, and A-type fingers. In the open-loop strategy, a high-level controller was used to actively adjust the motor inputs of the fingers, while a low-level controller executed these inputs by regulating both the agonist and antagonist tendons. This approach allows us to assess the system's mechanical intelligence based on its tendon-driven architecture. 

The operation involves executing various gestures by setting the desired setpoint on the encoder values of the agonist and antagonist motors. These setpoints range between When the Tactile SoftHand-A is in a fully open gesture and the DIP joint is not actively controlled (values 700 and 820, respectively) and fully closed (200 and 220, respectively). Active control of the DIP joint’s rotation can be achieved under any gesture when the encoder reading of the antagonist motor exceeds that of the agonist motor by approximately 200. When the encoder readings reach approximately 500, the DIP joint achieves its maximum rotational angle.

\subsection{Closed-loop Control Strategy}
The closed-loop control strategy integrates tactile and antagonistic tendon mechanisms to adjust the Tactile SoftHand-A's grasp gestures via tactile feedback, so as to prevent objects from slipping. Figure~\ref{fig_hand_control} provides a schematic diagram of the system. The system architecture employs a multi-processing approach, where a camera captures real-time images of the human hand, extracts joint landmark points using the python library MediaPipe, and calculates the angles between the fingers and palm. When the tactile sensors in the SoftHand-A fingertips do not detect contact, the SoftHand-A synchronizes with the human hand gestures by converting angles into motor inputs through a mapping function. Upon detection of contact, the system switches to closed-loop tactile feedback control and the Tactile SoftHand-A stops synchronizing human hand gestures. The tactile model then checks for slippage by determining if the displacement of the contact point exceeds a threshold, similarly to the method considered by~\cite{zhang2024compact}. If no slippage is detected, the SoftHand-A maintains the current grasp; conversely, if slippage is detected, the SoftHand-A adjusts the grasp gesture and rotates the DIP joints inward to apply a greater normal force, preventing the object from slipping. To release an object or grasp a new one, the human hand must be fully opened (with an average angle between the fingers and palm greater than 170 degrees) to exit the closed-loop tactile feedback control. The tactile SoftHand-A will then resynchronize with the human hand gestures until the next contact. A flow diagram of this process is given in the Appendix (Figure~\ref{fig_tactile_flow_chart}).

\section{Experiments} \label{experiments}

\subsection{Experiments A - Evaluation of Single Finger}\label{exp-finger}

\subsubsection{Experiment A1 - Comparison of Finger Gestures:}
This finger experiment investigates the respective controllability and capability to decouple the motion of the DIP and PIP joints of the D-type and P-type fingers, to show how coordinated control of the agonist and antagonist tendons allows active control of the appropriate joints. The A-type finger, which lacks the ability to decouple joints due to its antagonistic tendon arrangement, serves as a comparative model for the other types of fingers. For the D-type, P-type, and A-type fingers, Aruco marker codes affixed to the phalanges' sides track the angular changes of the DIP, PIP, and MCP joints. The motor inputs (encoder positions) are also taken and depicted as line graphs (Figure~\ref{fig_finger_gestures}).

\subsubsection{Experiment A2 - Dynamic Performance:}
This set of experiments examines the dynamic capabilities of the single fingers. Factors such as response speed, repeatable positioning accuracy, bearing capacity, and manipulation capability are assessed as described below.

A. {\em Response Time Test}: The response time of the finger during flexion and extension is evaluated between fully open or closed finger positions. Motor encoder values are used to monitor finger movements and the experiments videoed with a timer for extracting response times.

B. {\em Repeat Positioning Accuracy Test}: The finger’s ability to return to predefined positions is assessed with a laser rangefinder. This tracks the fingertip as it moves through 10 distinct positions from fully open to fully closed, with each position repeated over 10 trials.

C. {\em Bearing Capacity Test}: The finger’s load-bearing capability is evaluated under static conditions. Weights are applied to the medial phalange until there is greater than 10\,mm drop of the fingertip.

D. {\em Finger Manipulation Test}: The ability of the entire finger to push a stationary obstacle is assessed by finding the maximum weight moved by extending the finger. This movement is achieved by extending the finger. This experiment acts as a comparison for the next test.

E. {\em Fingertip Manipulation Test}: The fingertip's ability to manipulate objects through controlled movements is assessed by pushing a cuboid inward or outward from its ends and sliding it from above. These fine movements of the fingertip are achieved by coordinating the motion of the MCP and DIP joints. The maximal force exerted is measured.

In all cases, the experimental setup is depicted in the Appendix (Figures~\ref{fig_finger_supplement}A-E) and the results summarized in Table~\ref{tab:1}, with comparison against a single finger of the BPI SoftHand~(\cite{li2022brl}) that differs primarily in lacking the antagonist tendons and having just one motor.  

\subsection{Experiments B - Evaluation of Entire SoftHand}\label{exp-hand-gestures}
\subsubsection{Experiment B1 - Controllable Grasping Gestures:}
This hand experiment illustrates the capability of the Tactile SoftHand-A to perform controllable grasping gestures. Adjusting the input difference between the agonist and antagonist motors (Figure~\ref{fig_hand_gestures_results}A), can precisely control the grasping gestures, which is assessed by displaying both the front and side views under various inputs. 

\subsubsection{Experiment B2 - Hand Performance:}
This set of experiments assesses the SoftHand-A capabilities for reactive grasping and grasp strength. 

A. {\em Response Time Test}: The response time of the entire SoftHand during flexion and extension is evaluated between fully open or closed finger positions. The motor encoder values are used to monitor the finger movements, to demonstrate controllable speed and a smooth transition between agonist and antagonist motors over time. 

B. {\em Wrench Limit Test}: To assess the capability for safe and effective object grasping and manipulation by the SoftHand-A in various application scenarios, we used force sensors to measure the safe wrench limit of the robotic hand under stable grasping conditions.

C. {\em Grasp modes}: The SoftHand-A's capability to perform a power grasp, fingertip grasp and pinch grasp is compared experimentally to highlight the differences between these two modes when gripping objects. In particular, the pinch grasp experiment focuses on the coordination between the thumb and index finger, showcasing the effectiveness of this grasping style for tasks that require precision and control in handling smaller, thinner or more intricate objects. The results are shown in Figure~\ref{fig_fingertip_pinch_grasp}.

The experimental setups are depicted in the Appendix (Figure~\ref{fig_hand_response_wrench}A-B) and the results summarized in Table~\ref{tab:2}, with comparison against the BPI SoftHand~(\cite{li2022brl}) that differs in lacking the antagonist tendons and their motor.

\subsection{Experiment C - Grasping Adaptivity}\label{exp-adapt}
This experiment demonstrates the object-shape adaptivity of the hand. The Tactile SoftHand-A was mounted on the experimental bench controlled by the agonist and antagonist motors, enabling the Tactile SoftHand-A to grasp various objects (for the motor inputs shown in Figure~\ref{fig_hand_exp_adapt_results}(B)). We selected three regular objects: a triangular prism, a pyramid, and a cylinder; and two irregular objects: a wine glass and a tendon spool. The regular objects are of height 76\,mm, whereas the wine glass and tendon spool measure 136\,mm and 89\,mm in height, respectively.

This setup allowed evaluation of the SoftHand's adaptivity by comparing the contact conditions of each joint during the various grasps, with two types of grasp per object. In each case the object was placed manually within the hand, which was closed automatically onto the object.
To assess performance, we estimated the number of fingertips in contact, assessed from the tactile sensing, These contacts reflect the adaptive grasping ability of the robotic hand.

In addition, we compare the grasping ability of the Tactile SoftHand-A with that of the previous BPI SoftHand that has passive elements. The objects grasped include both YCB objects (\cite{calli2015ycb}) and common daily items. Grasp success number is calculated as the number of times the object is successfully grasped in 5 grasping experiments. A successful grasp occurs when no change in the contact area is detected within five seconds after the object is grasped under the influence of gravity; {\em i.e.}, no slippage occurs. The grasping data regarding the original BPI SoftHand is available from a previous study~(\cite{li2022brl}). The results of both experiments are shown in Figure~\ref{fig_hand_exp_adapt_results_comparison}.

\subsection{Experiment D - Tactile Feedback Control}\label{exp-d}
\subsubsection{Experiment D1 - Reaction to contact disturbances:}
To test the reactivity and sensitivity of the fingertip tactile sensing, we implemented a contact disturbance experiment (see Figure~\ref{fig_touch_follow}). In this experiment, the Tactile SoftHand-A made contact with an object, then the object's position is continuously disturbed through applying contact force, from which we assess that the hand can maintain contact. We utilize a pen as an indenter to apply arbitrary pressure to the fingertip.  The fraction of tactile deformation is estimated by 1 minus the structural similarity index (SSIM) of the indented tactile image from a tactile image of the undeformed state. This tactile deformation was calibrated against normal force in a separate experiment where a single finger is pressed against a force balance. We use a feedback controller that seeks to maintain the tactile deformation to a target setpoint, thereby ensuring the stability of the contact state. The reactivity of the hand can then be assessed from how the finger position moves dynamically in response to the real-time changes in the tactile image.

\subsubsection{Experiment D2 - Reactive Teleoperation:}
To evaluate the performance of our tactile feedback control system (shown in Figure~\ref{fig_hand_control}), this experiment uses teleoperation to adjust the grasp gestures using tactile feedback in response to slip. The experiment is structured into three sequential stages: synchronization, contact and slip. In the synchronization stage, the Tactile SoftHand-A aligns with the gestures of the human hand. During the contact and slip stages, the system engages in closed-loop tactile feedback control. In the slip stage, we assess the system's ability to detect slip and modify grasping gestures by artificially inducing slip to the thumb of the Tactile SoftHand-A, with a probe that we move across the fingertip. The tactile response of the hand uses the closed-loop control strategy described above. 

\subsubsection{Experiment D3 - Reaction to object disturbances:}
The above experiment is then extended to disturbances of objects held in hand. The experiment consists of three stages: synchronization and slip under external disturbance. During the synchronization stage, the BPI SoftHand-A mirrors human hand gestures to actively grasp objects. In the control group, grasping is performed entirely through teleoperation without tactile feedback, and once an object is successfully grasped, the hand maintains its grip. During the slip stage, the system switches to closed-loop tactile feedback control and stops synchronizing with human hand movements. The contact model detects slippage by determining whether the displacement of the contact point exceeds a predefined threshold. If slippage is detected, the Tactile SoftHand-A adjusts its grasp gesture and rotates the DIP joints inward to apply greater normal force, preventing the object from slipping. To quantify the effectiveness of these adaptive grasping strategies, we analyze the variations in normal force recorded by tactile sensors at the fingertips. This force-based evaluation provides quantitative evidence demonstrating how gesture adjustments contribute to improving grasp stability under dynamic conditions.

\begin{figure*}[t!]
		\centering
		\begin{tabular}[b]{c}
			\hspace{0cm}\includegraphics[width=1\textwidth,trim={0 0 0 0},clip]{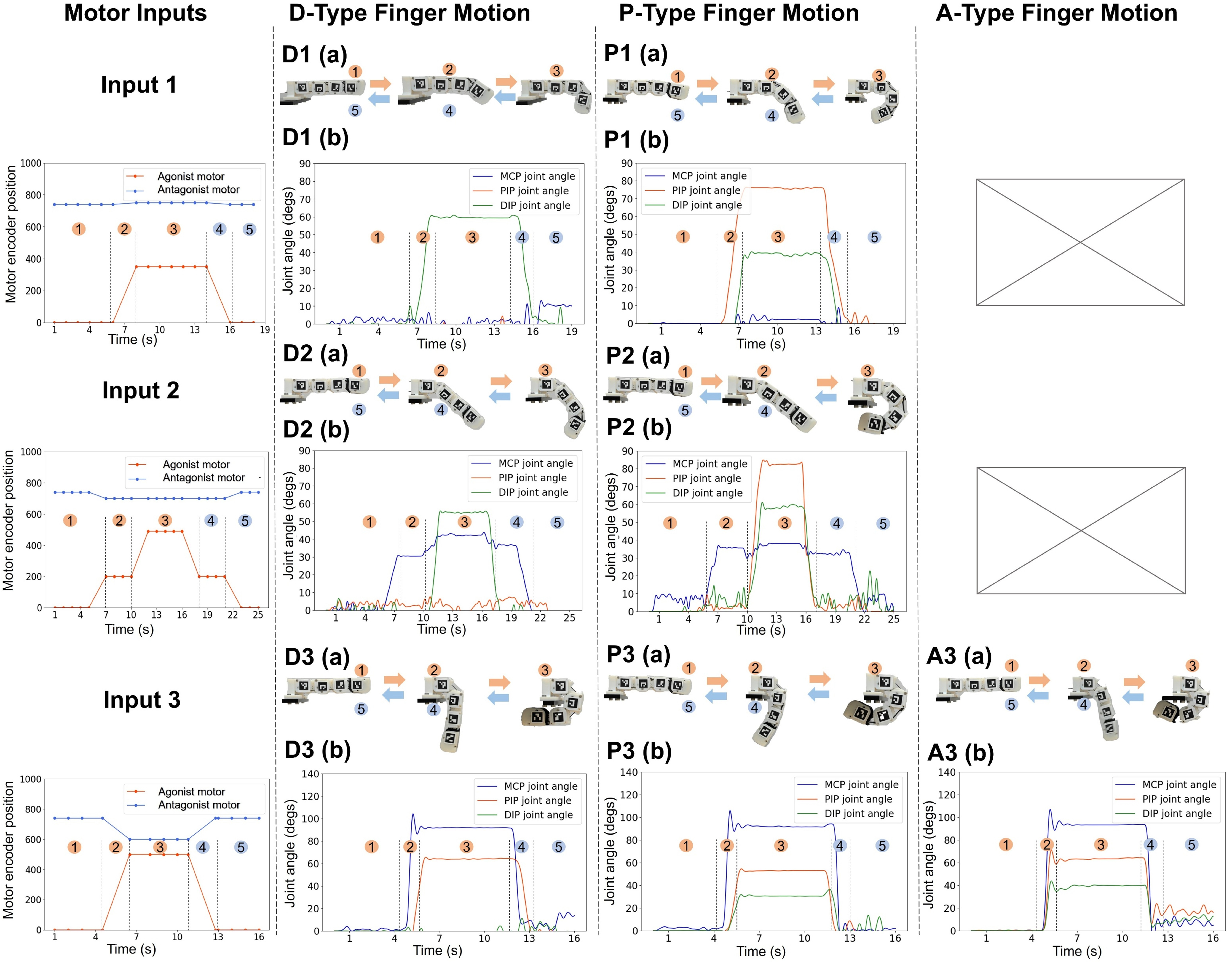} \\
		\end{tabular}
             \caption{Experiment A1: Motion of the D-type, P-type and A-type fingers under different control inputs to the agonist and antagonist motors (motor inputs 1 to 3, shown to left). The corresponding angles of the MCP, PIP and DIP joints in real time are shown to the right (plots D1-D3 for D-type; P1-P3 for P-type and just A3 for A-type). The A-type finger does not have the ability to move under inputs 1 and 2 due to the absence of soft elements, so there are no plots for A1 and A2. These plots show how it is possible to decouple the motion of the DIP or PIP joints for the D-type or P-type fingers.}
		\label{fig_finger_gestures}%
		\vspace{0em}
\end{figure*}
\begin{table*}[t!]
\resizebox{0.95\textwidth}{!}{%
	\renewcommand{\arraystretch}{1}
	\centering
	\begin{tabular}{@{}lcc@{}}	
		Test & SoftHand-A & BPI SoftHand \\
		\hline
		A1. Single finger response time closing from full extension to full flexion & 0.39\,sec & 0.66\,sec\\
        A2. Single finger response time opening from full flexion to full extension & 0.46\,sec & 0.32\,sec\\
        B. Single finger repeat positioning accuracy & Mainly within $\pm0.3$\,mm & Mainly within $\pm0.5$\,mm\\
        C. Single finger bearing capacity  & Drop under 8\,N load & Drop under 0.9\,N load\\
        D. Single finger manipulation: pushing & 7\,N & 2\,N \\
        E1. Fingertip manipulation: pushing & 3.5\,N & --\\
        E2. Fingertip manipulation: pulling & 9\,N & --\\
        E3. Fingertip manipulation: sliding & 1.2\,N & --\\
		\hline
	\end{tabular}}
	\caption{Experiment A2: Comparative single finger test results between the D-type finger of the SoftHand-A and the BPI SoftHand.}
	\label{tab:1}
\end{table*}

\begin{figure*}[t!]
		\centering
		\begin{tabular}[b]{c}
	\hspace{0cm}\includegraphics[width=0.95\textwidth,trim={0 0 0 0},clip]{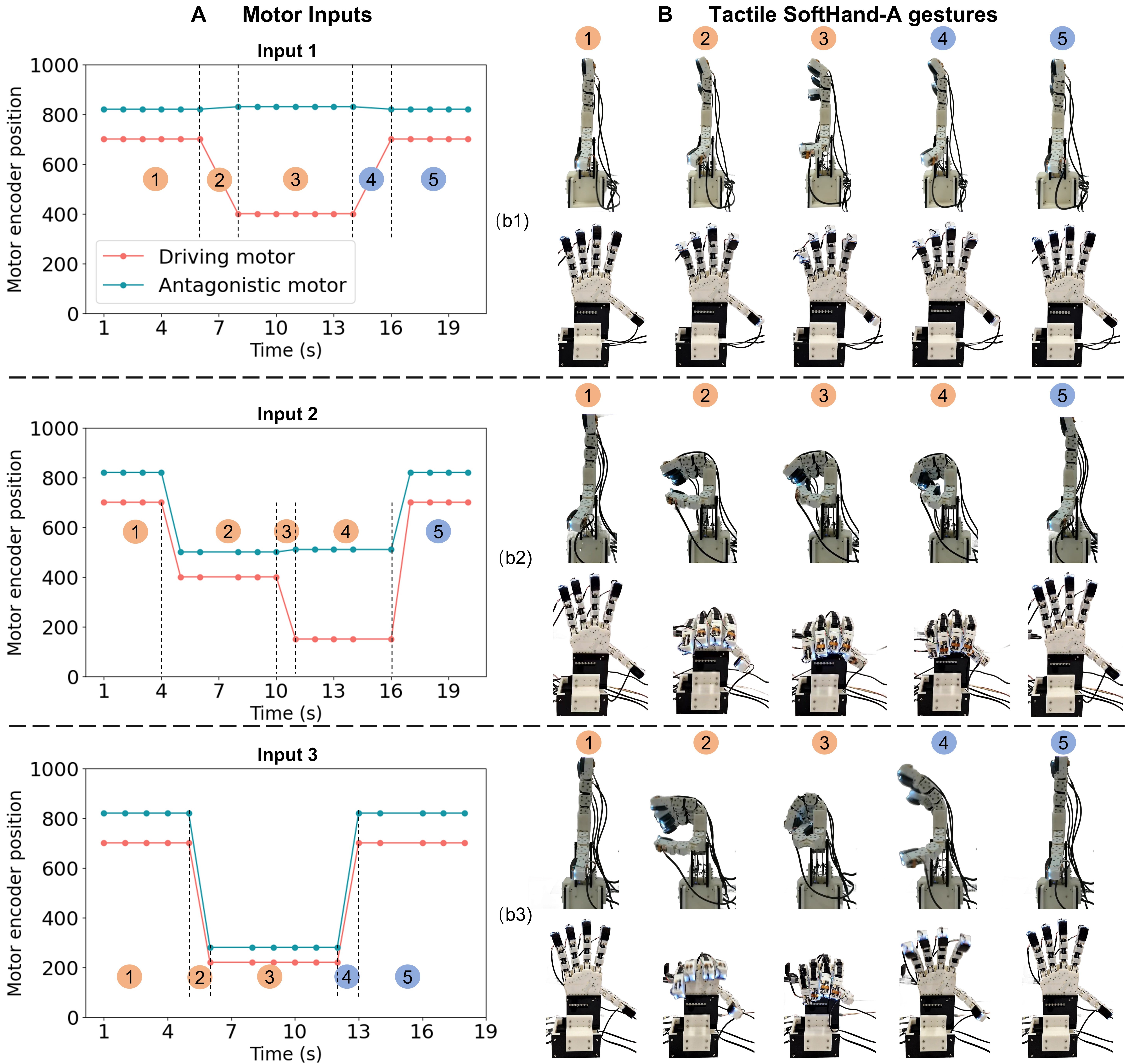} \\
		\end{tabular}
             \captionof{figure}{Experiment B1: Response of the entire SoftHand-A (D-type fingers) to motor inputs. (A) The three example motor inputs are intended to show that the SoftHand-A can actively control its open/closure and the DIP joints through the coordinated control of the agonist and antagonist motors. (B) SoftHand-A gesture changes under these inputs at different times in the trajectory. Red points indicate the closing trajectory and the blue points show the opening trajectory of the SoftHand-A.}
		\label{fig_hand_gestures_results}
\vspace{2em}
\resizebox{0.92\textwidth}{!}{%
	\renewcommand{\arraystretch}{1}
	\centering
	\begin{tabular}{@{}lcc@{}}	
		Test & SoftHand-A & BPI SoftHand \\
		\hline
		A1. Hand closure response time from full extension to full flexion & 0.46\,sec & 1.38\,sec\\
        A2. Hand opening response time from full flexion to full extension & 0.59\,sec & 0.96\,sec\\
        B3. Maximum wrench forces (in $x$, $y$, $z$) & 20\,N, 8\,N, 8\,N & 11.4\,N, 4.8\,N, 3.5\,N\\
        B3. Maximum wrench torques (around $x$, $y$, $z$) & 0.6\,Nm, 0.3\,Nm, 0.6\,Nm & 0.23\,Nm, 0.24\,Nm, 0.5\,Nm\\
        C. Grasp modes & Power, Fingertip, Pinch & Power\\
        \hline
	\end{tabular}}
	\captionof{table}{Experiment B2: Comparative whole hand test results between the SoftHand-A and the BPI SoftHand.}
	\label{tab:2}
\end{figure*}

\begin{figure*}[t!]
		\centering
		\begin{tabular}[b]{c}
	   \hspace{0cm}\includegraphics[width=0.9\textwidth,trim={0 0 0 0},clip]{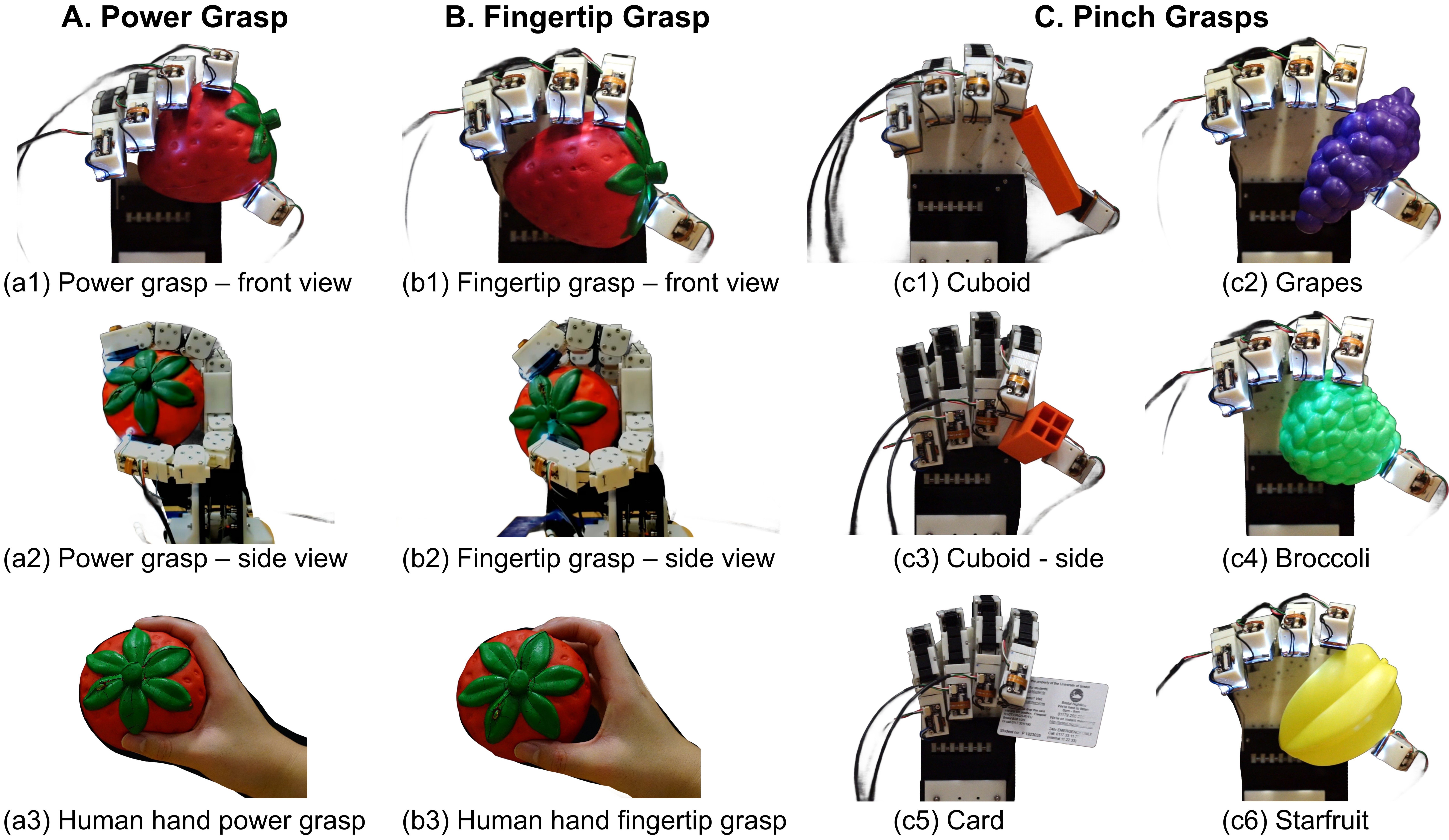} \\
		\end{tabular}
             \caption{Comparison between power grasp (A) fingertip grasp (B) and pinch grasps (C). The difference between the power and fingertip grasp are shown with the two different grasping modes of human hand (a3,b3). Front (a1,b1) and side (a2,b2) views are shown for the power and fingertip grasps. Pinch grasp results are shown with a range of objects held in different ways (c1-c6).}
		\label{fig_fingertip_pinch_grasp}%
		\vspace{-0em}
\end{figure*}

\section{Results}\label{Results}
\subsection{Experiment A Results}
\subsubsection{Results A1 - Comparison of Finger Gestures:}
First, we use the D-type (distal-type) finger to illustrate the decoupled control of the DIP joint through coordinated control of the agonist and antagonist motors. Initially, in the finger extension state, the agonist motor is position-controlled to rotate a specific angle, pulling the agonist tendon and generating a closing force on the finger. The antagonist motor can be position controlled to modify the opposing tension of the antagonist tendon, giving different gestures of the finger (motor inputs 1, 2 and 3 in Figure~\ref{fig_finger_gestures}), and also to extend the finger to actively return it to an open posture.

The first motor input flexes the finger under the agonist motor (input 1 in Figure~\ref{fig_finger_gestures}) while allowing a large differential with the antagonist motor. The D-type and P-type fingers differ in that the D-type can be controlled to flex just the DIP joint due to opposing forces at the PIP and MCP joints, unlike the P-type where the MCP and PIP joints are also flexed. This process is reversed to extend the finger.

The second motor input has two initial stages, starting with a small flexion of the finger in which the active antagonism is reversed to lower the antagonist tension, then a larger drive flexes the entire finger (input 2 in Figure~\ref{fig_finger_gestures}). The D-type and P-type fingers behave similarly in flexing the entire finger, except the P-type finger has a larger flexion. Again, this process is reversed to extend the finger.

The third motor input is similar to the first input to flex the finger, but has a larger de-tensioning of the antagonist tendons (input 3 in Figure~\ref{fig_finger_gestures}). All three D-type, P-type and A-type fingers behave similarly in flexing all joints.

\begin{figure*}[t!]
		\centering
		\begin{tabular}[b]{c}
	\hspace{0cm}\includegraphics[width=1\textwidth,trim={0 0 0 0},clip]{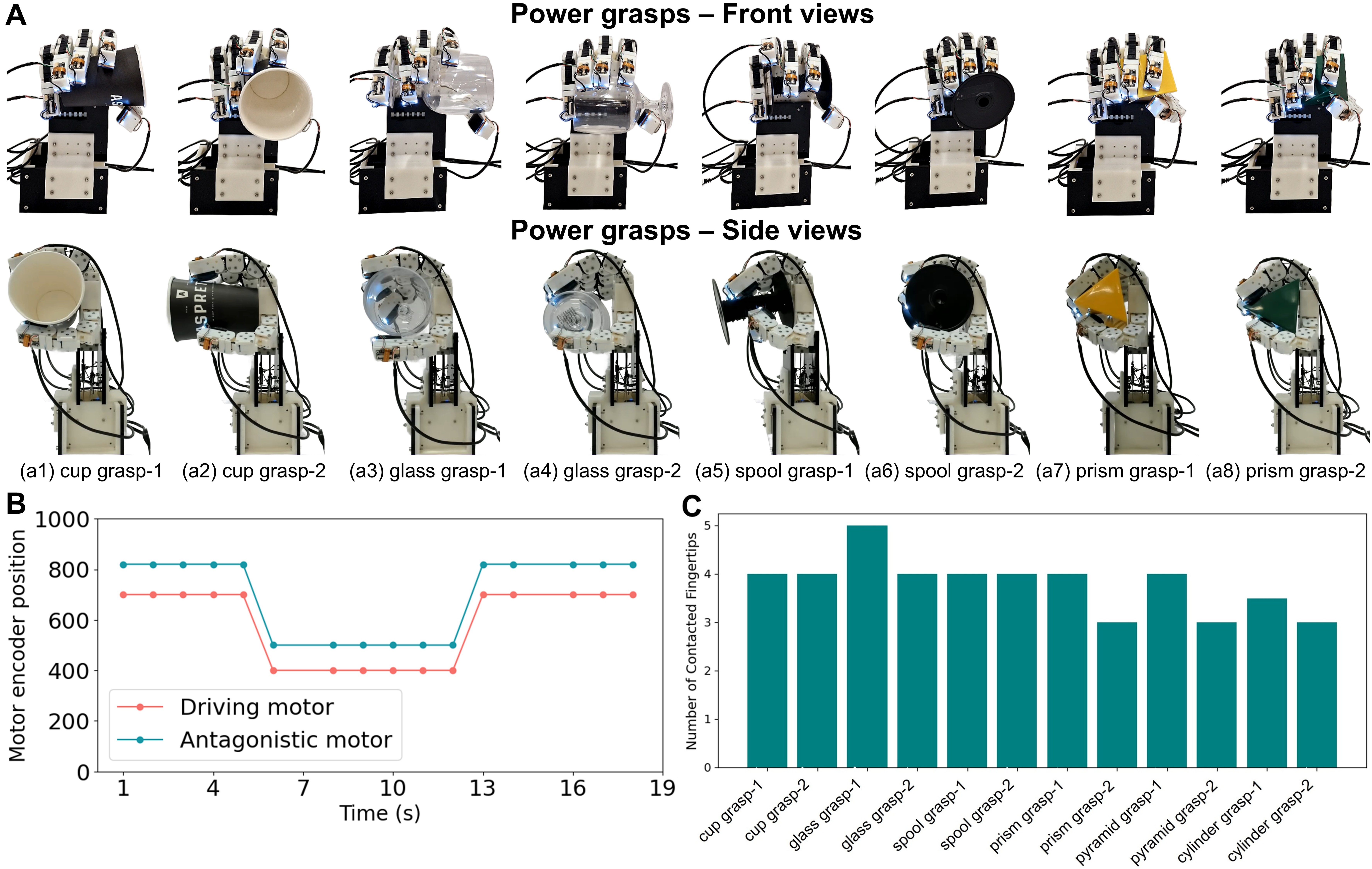} \\
		\end{tabular}
        \vspace{-1.5em}
             \caption{Grasping adaptivity of the tactile SoftHand-A with D-type fingers to various objects with the same motor inputs. (A)~Front and side views of the hand at maximal closure. (B) Input from the agonist and antagonist motors. (C) The number of contacting fingertips at maximum closure (validated from the tactile outputs).}
		\label{fig_hand_exp_adapt_results}%
		\vspace{0em}
		\centering
		\begin{tabular}[b]{c}
            \includegraphics[width=0.95\textwidth,trim={0 0 0 0},clip]{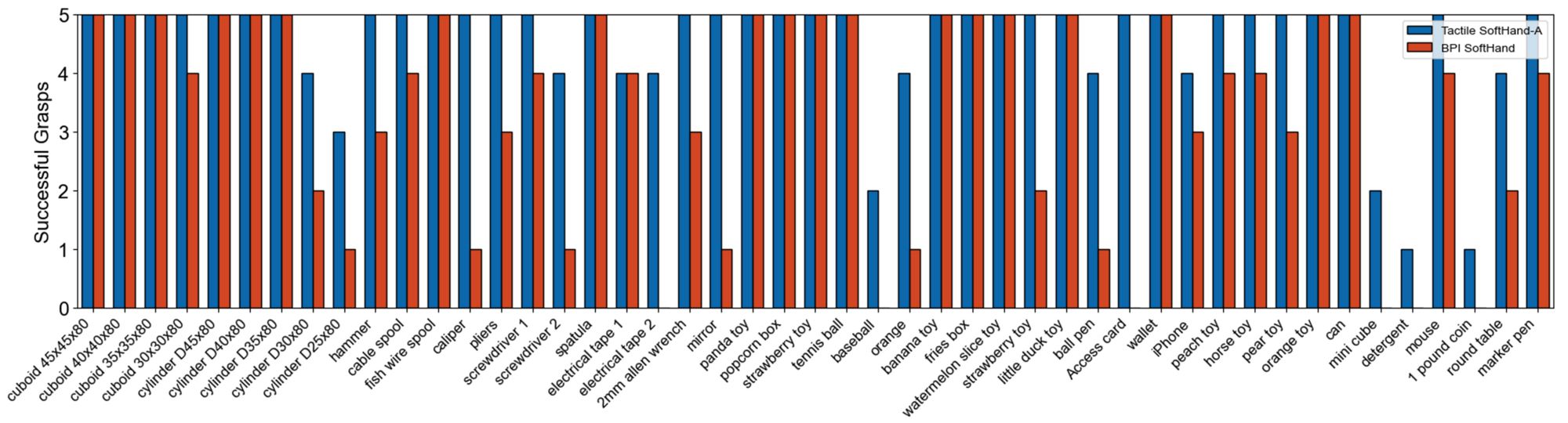} \\
            \vspace{-2em}
		\end{tabular}
             \caption{Comparison of effective grasping with with Tactile SoftHand-A (shown in blue) and the BPI SoftHand with a passive reset (shown in red) incorporating passive elements. Results for the BPI SoftHand are from \cite{li2022brl}.}
		\label{fig_hand_exp_adapt_results_comparison}%
		\vspace{0em}
\end{figure*}

\subsubsection{Results A2 - Dynamic Performance:}
Multiple performance measures for a single D-type finger of the SoftHand-A are summarized in Table~\ref{tab:1} (with experiments shown in Appendix Figure~\ref{fig_finger_supplement}). These tests assessed properties such as response speed, repeatable positioning accuracy, bearing capacity, and manipulation capability. In all cases, the results are compared with a single finger of the BPI SoftHand~(\cite{li2022brl}) to quantify the performance improvements from active antagonism.

Overall, the D-type finger of the SoftHand-A is able to rapidly flex and extend, taking less than 0.5\,sec between fully open and closed positions (corresponding to joint motions of 2-3\,rad/sec). In contrast, the finger of the BPI SoftHand is slower to flex to full closure (by 0.17\,sec), but can more rapidly extend to opening (by 0.14\,sec) due to the passive elastic reset that is activated immediately when tension is removed from the agonist motor.

Likewise, the repeat positioning is more accurate for the D-type finger of the SoftHand-A (within $\pm0.3$\,mm) than for a finger of the BPI SoftHand (within $\pm0.5$\,mm). This is attributed to the more precise control from having both agonist and antagonist motors.

The single finger bearing capacity is much higher for the SoftHand-A finger (8\,N) compared tothe BPI SoftHand finger (0.9\,N). This is expected because the SoftHand-A is able to bear load by tensioning the antagonist tendon, but the BPI SoftHand only has a passive elastic mechanism to maintain finger extension.

The single finger manipulation experiments compared simple actions such as pushing, pulling and sliding from above (Appendix Figure~\ref{fig_finger_supplement}D,E). For a baseline, pushing with the entire finger was considered, with a maximum load of 7\,N for the SoftHand-A that was 5\,N larger than that for the BPI SoftHand. Finally, only the SoftHand-A could demonstrate fingertip object manipulation, as the position of its DIP joint is controllable (see previous section on grasp gestures). The experimental results showed that by utilizing the two-tendon system, the DIP joint could be actively controlled while simultaneously reactivating the antagonist tendon to rotate the MCP joint, providing sufficient workspace for the fingertip to manipulate the object by pushing, pulling and sliding.

\subsection{Experiment B Results}
\subsubsection{Results B1 - Controllable Grasping Gestures:}
The SoftHand-A uses the D-type joint so that control over the DIP joints of all fingers can be achieved by increasing the differential between the agonist motor and the antagonist motor. Initially, activating the antagonist motor tightens the antagonist tendon, thereby preventing the fingers' PIP and MCP joints from bending actively; then its direction can be reversed in phase 3 to change the DIP joint angles. The experiments show the front and side views of the SoftHand-A under three motor inputs analogous to those used to assess single finger gestures (Figures~\ref{fig_hand_gestures_results}). 

The first motor input flexes the fingertips using the agonist motor with a large differential with the antagonist motor (input 1 in Figure~\ref{fig_hand_gestures_results}). This motor activation flexes just the DIP joints due to opposing forces at the PIP and MCP joints. The reverse process extends the fingertips by rotating just the DIP joints.

The second motor input has an initial stage that starts with a small flexion of the finger in which the active antagonism is reversed to lower the antagonist tension, followed by a larger drive that closes the entire hand (input 2 in Figure~\ref{fig_hand_gestures_results}). The initial stage then enables active control of the DIP joints while the hand is in a semi-closed state (Figure~\ref{fig_hand_gestures_results}, middle panels). The hand is opened by activating the agonist and antagonist motors to return to their initial positions.

The third motor input is similar to the first input to flex the finger, but drives the antagonist motor to decrease the differential between the two motors, de-tensioning the antagonist tendons (input 3 in Figure~\ref{fig_hand_gestures_results}). This motion also results in a grasp gesture that closes the entire hand, but without the active control of just the DIP joints. The hand is opened by returning the motors to their initial positions.

\begin{figure*}[t!]
		\centering
		\begin{tabular}[b]{c}
	\includegraphics[width=1\textwidth,trim={0 0 0 0},clip]{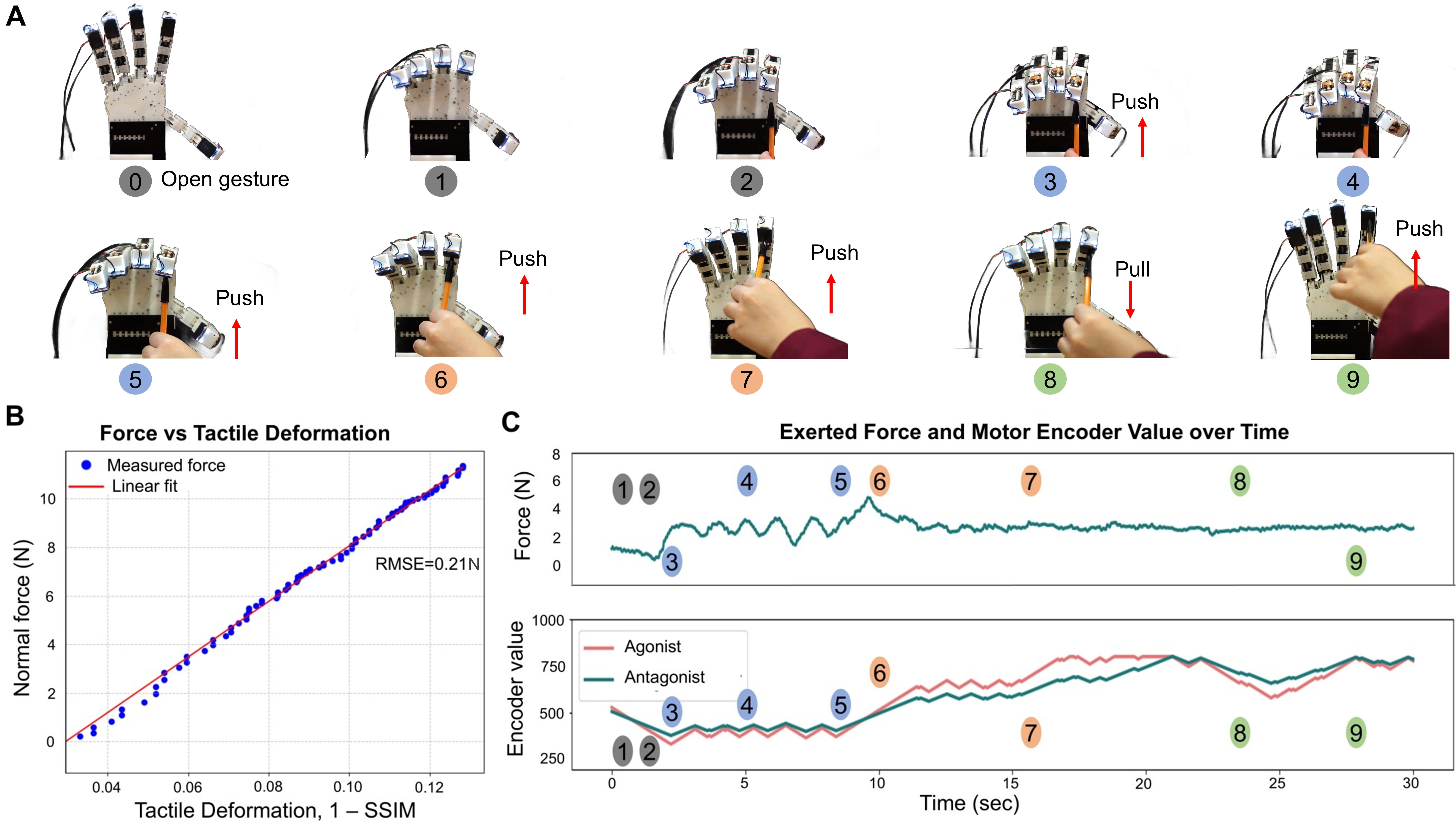} \\
		\end{tabular}
             \caption{Reaction to contact disturbances with the Tactile SoftHand-A. (A) Interactions between Tactile SoftHand-A and indenter, showing 10 stages where the hand closes onto the indenter, the the fingertip is pushed, pulled (by withdrawing the indenter) and pushed back again. The red arrows show the direction that the indenter is moving. (B) Calibration of tactile deformation (1-SSIM) with normal force measured from a strain gauge, where SSIM measures the similarity of the tactile image with a reference undeformed image. (C) Top, change in inferred force over time for the tactile fingertip of the index finger; bottom: change in encoder values over time for the agonist and antagonist motors.}
		\label{fig_touch_follow}%
\end{figure*}

\begin{figure*}[t!]
		\centering
		\begin{tabular}[b]{c}
	\hspace{0cm}\includegraphics[width=1\textwidth,trim={0 0 0 0},clip]{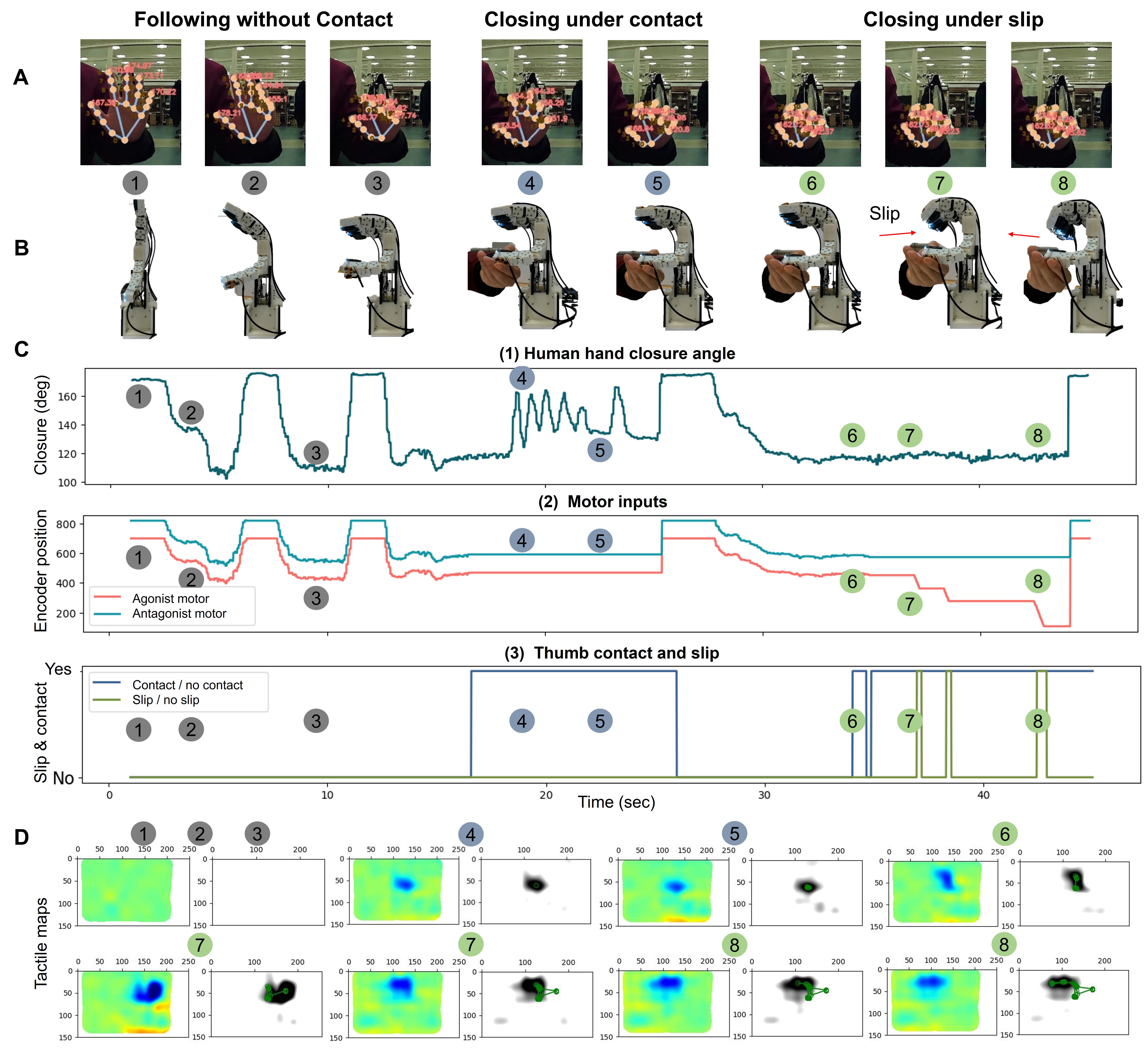} \\
		\end{tabular}
             \caption{Human hand gesture and tactile control of the Tactile SoftHand-A. (A) human hand gestures used as inputs to control the hand. (B) Side view of Tactile SoftHand-A in response to the gesture control. (C-1) Average degree of human finger closure, (C-2) corresponding motor inputs to the hand, (C-3) and change of finger contact and slip. (D) Tactile contact map at various stages throughout the motion as the hand contacts and object. The results in this figure relate to Experiment D.}
		\label{fig_hand_tac_grasping_results}%
		\vspace{-0em}
\end{figure*}

\begin{figure*}[t!]
        \begin{tabular}[b]{c}
	\hspace{0cm}\includegraphics[width=1\textwidth,trim={6 0 0 0},clip]{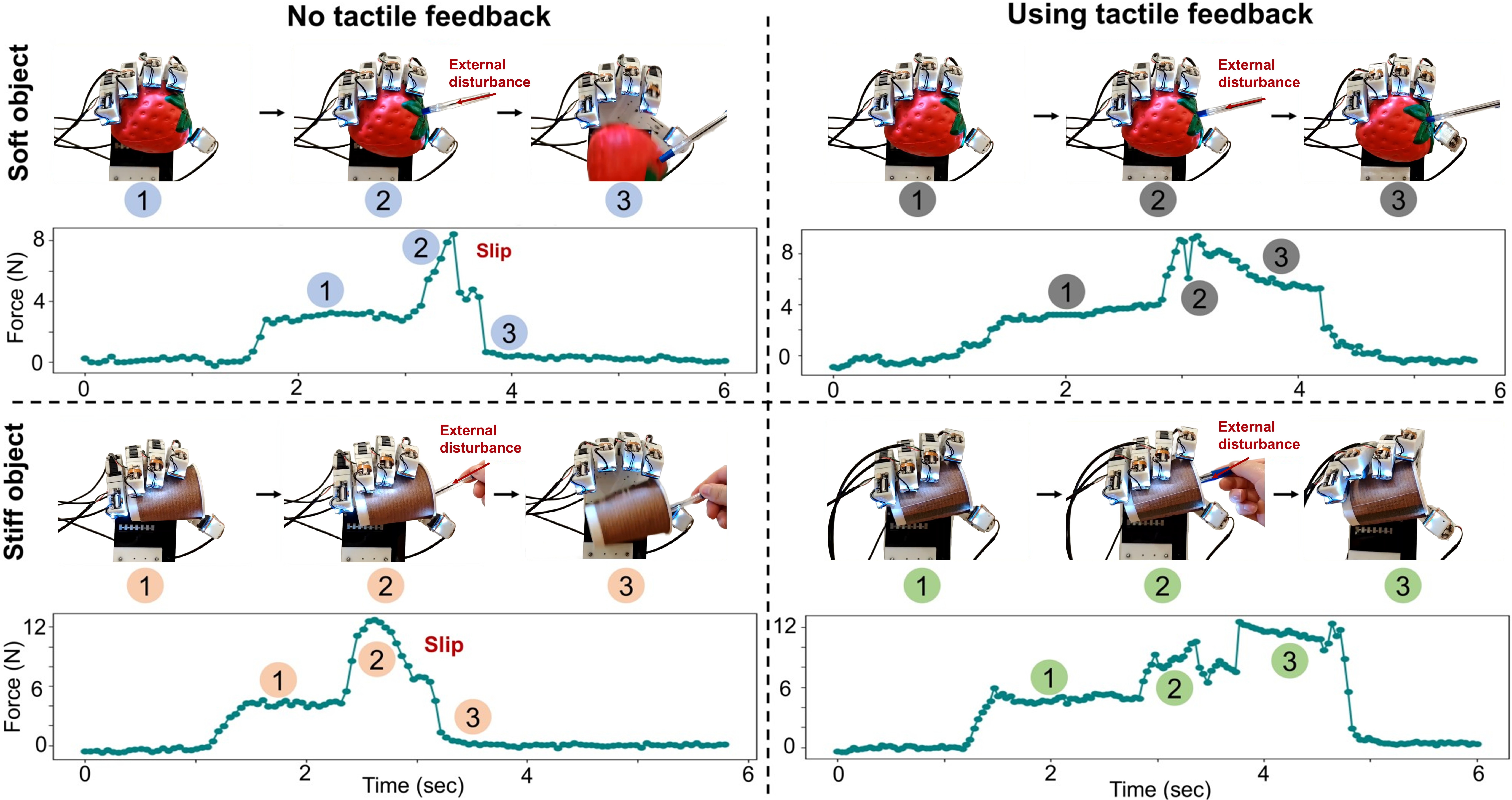} \\
		\end{tabular}
             \caption{Comparison of grasping under an external disturbance with and without tactile feedback (left and right panels, respectively). Phase 1: a teleoperator closes the hand to stably hold the object. Phase 2: an external disturbance is applied to the held object. Phase 3: with no tactile feedback the hand is unable to respond to this disturbance, otherwise with tactile feedback the hand can tighten its grip to prevent the object being dropped.}
		\label{fig_hand_tac_stable_grasping}%
		\vspace{-0em}
\end{figure*}

\subsubsection{Results B2 - Hand Performance:}
The hand performance evaluation is summarized in Table~\ref{tab:2} and experiments depicted in the Appendix (Figure~\ref{fig_hand_response_wrench}). Overall, the hand demonstrated rapid movements of $\sim 0.5$\,sec between fully open and closed positions with a maximum normal wrench force of 20\,N and 9\,N in the other directions, and maximum wrench torques of $0.3$\,Nm to $0.6$\,Nm.

A more detailed depiction of the wrench forces and torques over time is shown in the Appendix (Figure~\ref{fig_hand_response_wrench}(B)). The forces $(F_x, F_y, F_z)$ are applied within the coordinate frame of the hand. We note that the normal force $F_z$ peaks at around 20\,N. Similarly, the maximal torque is in the axis $R_z$ and reaches a peak value of 0.6\,Nm. These plots illustrate the robotic hand's capacity to apply significant forces and torques, indicating its capability for physical interaction during various tasks.

The SoftHand-A is able to effectively form power, fingertip and pinch grips on various objects~(Figure~\ref{fig_fingertip_pinch_grasp}). When a power grasp is employed, the fingers conform to the surface of the object, increasing the contact area and allowing for a secure grasp (Figure~\ref{fig_fingertip_pinch_grasp}A), which is particularly suitable for grasping heavier objects. For delicate objects with easily damaged surfaces, such as fruits, the SoftHand-A can switch to the Fingertip Grasp mode (Figure~\ref{fig_fingertip_pinch_grasp}B), like the human hand. In this mode, only the soft fingertips make contact with the object's surface, effectively avoiding the potential for surface damage that may occur in power grasp mode due to the uncontrolled rotation or pressure across the hand. Finally, the SoftHand-A is also able to grasp a range of small or thin objects with pinch grasps between the thump and forefinger (Figure~\ref{fig_fingertip_pinch_grasp}c). This flexibility in grasping modes ensures that the Tactile SoftHand-A can handle a wide variety of objects while minimizing the risk of damage to sensitive items.

\subsection{Results C - Grasping Adaptivity}
The interactions between the Tactile SoftHand-A and various grasped objects are shown in Figures~\ref{fig_hand_exp_adapt_results}A. Specifically, the thumb, positioned opposite the other four fingers, forms an arc to encircle and stabilize objects. This arrangement and differential mechanism allow the remaining fingers to flex to various extents, adapting to the object's shape, which demonstrates the Tactile SoftHand-A's adaptivity to different object geometries. 

For many of the objects, the little finger (pinkie) to retracts towards the palm, giving additional enclosure, as is most evident in Figures~\mbox{\ref{fig_hand_exp_adapt_results}a5-a8}. This adaptivity is attributed to the novel differential mechanism, which is not limited to the little finger alone, and is also evident in the variation in the number of contracted fingertips~(Figure~\ref{fig_hand_exp_adapt_results}C).  

For the irregular objects, each finger joint adjusts its bending to conform to the object's shape (see Figures~\ref{fig_hand_exp_adapt_results}a1-a6), showcasing the SoftHand-A's adaptability in grasping varied geometries. For example, after the little finger, thumb, and index finger of the Tactile SoftHand-A make contact with the outer flange of the tendon spool, the ring and middle fingers can still bend until they touch the central axis.

In Figure~\ref{fig_hand_exp_adapt_results}(D), we compared the grasping ability of the Tactile SoftHand-A with that of the BPI SoftHand incorporating passive elements (\cite{li2022brl}). For most objects, the Tactile SoftHand-A demonstrated superior or equivalent performance, successfully grasping many items in five out of five attempts. Objects such as cuboids, cylinders, cable spools, and several soft toys ({\em e.g.}, a strawberry toy, popcorn toy, and orange toy) consistently exhibited higher grasp success rates, with the Tactile SoftHand-A achieving the maximum possible success over five grasps. For some smaller or irregularly shaped objects, such as a ball, pen and coins, or heavier objects like detergent, the success rate dropped for both hands, indicating the difficulty these hands face with these items. However, the Tactile SoftHand-A generally maintained higher consistency and adaptivity across a wider range of tested objects, especially with everyday tools and toys. This comparison highlights the enhanced robustness and effectiveness of the Tactile SoftHand-A in handling diverse shapes and sizes, particularly in scenarios requiring more precise or adaptive grasping strategies.

\subsection{Results D - Tactile Feedback Control}
\subsubsection{Results D1 - Reaction to Contact Disturbances:}
Various responses of the Tactile SoftHand-A during changes in contact are shown in Figure~\ref{fig_touch_follow}(A). Throughout the experiment, the contact (here a tip of a pen) underwent a series of actions implemented manually, such as being pushed forward or retracted (phases 0-9). In the initial phases (phases 1-2), the tactile fingertips were not yet in contact with the pen, and the Tactile SoftHand-A closed its fingers until contact was made in phase 3. Subsequently, the pen was manually moved upward, resulting in increased pressure on the tactile fingertips and the Tactile SoftHand-A responded by extending its fingers to maintain a target deformation and preserve a stable contact. In phases 6-7, as the pen continued to move upward, the fingers extended further in response to this motion. Then in phase 8, when the pen began to retract, the fingers instead flexed to maintain the deformation.

As described in the experimental sections, the control law seeks to maintain a fraction of deformation value at a target of 0.05, quantified by the structural similarity index measure (SSIM) of the current tactile image from an undeformed tactile image. A calibration of 1-SSIM against normal force (Figure~\ref{fig_touch_follow}(B)) revealed a near-linear relation, such that the target corresponds to a normal force of approximately 2\,N. Consequently, the control law seeks to adjust the motor encoder values to maintain a normal force of this value, which is evident in the plots of these quantities over time (Figure~\ref{fig_touch_follow}(C)). 

These experimental results demonstrate that the Tactile SoftHand-A can use its tactile feedback to effectively maintain a stable contact during dynamic tasks.

\subsubsection{Results D2 - Reactive Teleoperation:}
This experiment is structured into three sequential stages: synchronization, contact, and slip. In the synchronization stage, the Tactile SoftHand-A aligns with the gestures of the human hand. During the contact and slip stages, the system engages in closed-loop tactile feedback control. The results are shown in Figure~\ref{fig_hand_tac_grasping_results}.

Various time points during the gesture synchronization phase (1-15 seconds), are shown in Figure~\ref{fig_hand_tac_grasping_results} from time points 1-3. At this stage, Tactile SoftHand-A synchronizes with the gestures of the human hand. For instance, at time point 1, Figure~\ref{fig_hand_tac_grasping_results}(A) illustrates the human hand in an open position, while Figure~\ref{fig_hand_tac_grasping_results}(B) shows the Tactile SoftHand-A in a similar fully open state. By time point 3, Figure~\ref{fig_hand_tac_grasping_results}(A) depicts the human hand partially closed, and the Tactile SoftHand-A mirroring this gesture in Figure~\ref{fig_hand_tac_grasping_results}(B). Figure~\ref{fig_hand_tac_grasping_results}(C)(2) demonstrates that the input curves of the two motors follow a similar trend to the finger angle curve of the human hand, showcasing effective synchronization of the human and robot hand gestures. Notably, the delay between these curves is less than one second, attributable to the anti-tendon mechanism in Tactile SoftHand-A, which facilitates rapid response. 

From approximately 16 to 26 seconds, the contact stage occurs, during which the thumb's tactile fingertip of Tactile SoftHand-A makes contact with the human hand. Concurrently, contact detection is indicated in Figure~\ref{fig_hand_tac_grasping_results}(C)(3), where the blue curve representing the thumb slip/contact state remains in contact throughout this stage. Additionally, Figure~\ref{fig_hand_tac_grasping_results}(D), images 4-5, depict the contact area and the location of its center point (green circle marker). Furthermore, during this stage, the angle of the human hand’s fingers (shown in Figure~\ref{fig_hand_tac_grasping_results}(C)(1) is deliberately altered to introduce interference, yet the inputs to the two motors are kept constant to preserve the current grasp of Tactile SoftHand-A. This indicates that the control system has engaged in closed-loop tactile feedback control, unaffected by changes in the human hand’s finger angles, thereby enhancing grip stability. The constant gestures of Tactile SoftHand-A are shown in Figure~\ref{fig_hand_tac_grasping_results}(A)(4-5).

From approximately 26 to 33 seconds, the gesture synchronization phase occurs, during which Tactile SoftHand-A is controlled by the human hand to re-establish contact. From 33 to 43 seconds, the slip phase takes place. During this phase, Figure~\ref{fig_hand_tac_grasping_results}(D)(7-8) illustrates the changes in the contact area and the slip of its center point, clearly showing the contact point moving first to the right and then slipping to the left. The green curve in Figure~\ref{fig_hand_tac_grasping_results}(C)(3), which represents slip detection, also remains in the slip state during this period, indicating that slippage has occurred. At the same time, the contact is also indicated (blue curve). Following the onset of slippage, the motor inputs change as shown in moments 7 and 8 of Figure~\ref{fig_hand_tac_grasping_results}(C)(2), adjusting the grip gesture of Tactile SoftHand-A by increasing the input differential between the agonist motor and the antagonist motor. As demonstrated in Figure~\ref{fig_hand_tac_grasping_results}(B)(7-8), the DIP joints of Tactile SoftHand-A's fingers actively rotate inward, adjusting their grip to enhance the gripping force and prevent the object from slipping.

\subsubsection{Results D3 - Reaction to Object Disturbances:}

Our final experiment extends the teleoperated control to grasping performance when external disturbances are applied to the held object, comparing the influence with and without tactile feedback (see Figure~\ref{fig_hand_tac_stable_grasping}). 

In the non-tactile feedback experiments, the SoftHand-A first successfully grasped a soft toy (shaped like a strawberry) using a teleoperation procedure (denoted Phase 1), with inferred fingertip contact forces stabilizing around 3.5\,N. When external disturbances were applied (Phase 2), the inferred contact force rapidly increased to $\sim$8\,N, followed by an abrupt decline to 0\,N (denoted Phase 3) due to object slippage (Figure~\ref{fig_hand_tac_stable_grasping}A). The entire experiment was repeated with a stiffer object (a cup), which equilibrated to a higher contact force of 12\,N, before likewise failing to securely hold the object under an external disturbance (Figure~\ref{fig_hand_tac_stable_grasping}C).

In contrast, the experiments with tactile feedback exhibited successful reactions to external disturbances, as the hand could react autonomously to stabilize the object. During Phases 1 and 2, the inferred contact forces remained consistent with the non-tactile experiments. However, phase 3 was significantly different: upon detecting a change in contact force induced by disturbances, the control system triggered a compensation mechanism that adjusted just the DIP joint to bring the fingertips into greater contact with the object. While maintaining object retention and counteracting dynamic external disturbances, the Tactile SoftHand-A achieved a reduction of contact forces, ultimately establishing an equilibrium between the disturbance components of the force and the other contact forces to achieve a stable grasp. This tactile feedback successfully prevented secondary object slippage, in contrast to the object loss observed without tactile feedback. 

These comparative experiments demonstrate the adaptive control capabilities of using tactile feedback coupled with the antagonistic tendon mechanism of the tactile SoftHand-A.

\section{Discussion}
In this paper, we have presented and tested our design of the Tactile SoftHand-A, a 3D-printed highly-underactuated, anthropomorphic robot hand with an antagonist tendon mechanism based on the BPI-SoftHand~(\cite{li2022brl}), a 3D-printed derivative of the Pisa/IIT SoftHand~(\cite{catalano2014adaptive}). The SoftHand-A has improved dexterity over the original purely agonist mechanisms with passive reset through elastic elements. The Tactile SoftHand-A also features vision-based tactile sensors that are completely 3D-printed within the hand structure, using the TacTip design of marker-based tactile sensors~(\cite{ward2018tactip,lepora2021soft}). The experimental results assessed the novel control capabilities possible with the Tactile SoftHand-A. The key findings from these results are discussed below.

First, we tested the performance of a single finger, considering three versions differing in the termination of the antagonist tendon: the A-type finger terminating at the tip of the distal phalange, the D-type finger on the medial phalange, and the P-type on the proximal phalange (Figure~\ref{fig_finger_gestures}). Of these, the A-type finger is the most reactive; however, the D-type finger permits control of the distal interphalangeal (DIP) joint. By managing the position and tension of the tendons, suitable forces are generated at the PIP and MCP joints to allow precise control of the DIP joint. This level of control enables tasks requiring fine motor skills, such as changing the hand gesture to prevent slippage while not appreciably increasing the grasping force. Because of this versatility, this design is adopted in the SoftHand-A.

Quantitative comparisons between a single finger of the SoftHand-A (D-type) and the BIP SoftHand showed several benefits of this design. Overall, the antagonistic mechanism results in a moderate improvement in reaction speed and accuracy, and a large increase in weight bearing capacity~(Table~\ref{tab:1}). Overall, the most significant improvements are in manipulation capabilities such as pushing, pulling, and sliding, which the precursor BPI SoftHand is unable to perform. 

Next, we tested the whole-hand performance of the SoftHand-A, considering the version with the D-type finger due to its improved manipulation capabilities. By adjusting the agonist and antagonist motors, the DIP joints of all fingers can be controlled, as shown in Figure~\ref{fig_hand_gestures_results}. This method prevents the PIP and MCP joints from active bending, allowing for isolated control of the DIP joints, which is needed for tasks requiring a stable grip while maintaining finger dexterity. As a consequence, the Tactile SoftHand-A is capable of performing various grasping modes, including power grasp, fingertip grasp, and pinch grasp that all benefit from the hand's adaptivity onto various object shapes (Figure~\ref{fig_fingertip_pinch_grasp}). The fingertip grasp is a new grasp enabled by the D-type finger that is suitable for handling delicate objects, where reduced contact between the hand and the object protects a fragile surface while maintaining a stable grip.  By switching between these different modes, the SoftHand-A can flexibly and adaptively adjust its grasping strategy based on the characteristics of the object.

A quantitative comparison of the Tactile SoftHand-A and the BPI SoftHand shows significantly improved reaction times and wrench forces/torques due to the antagonistic tendon mechanism (Table~\ref{tab:2}). Improvements are around a factor of two, with the closure time differing most (reducing from 1.4\,sec to 0.5\,sec). These originate from not having to work against a restoring force that the BPI SoftHand uses to passively reset to an open position.  

The adaptivity of the Tactile SoftHand-A is further evidenced in grasping experiments with various objects (Figure~\ref{fig_hand_exp_adapt_results}). The thumb's positioning opposite the other fingers allows for effective encircling and stabilization of objects, enhancing the grip's support. The differential mechanism enables each finger to adjust its flexion based on the object's shape, which is particularly important for robot hands that must interact with a variety of objects in daily life. Likewise, a comparison with the BPI SoftHand demonstrated a significantly improved adaptive capability to achieve secure grasps~(Figure~\ref{fig_hand_exp_adapt_results_comparison}).

Then three experiments test the sensing capabilities of the Tactile SoftHand-A and the reactive control that the tactile sensing gives. First, we test the reactivity to contact disturbances through real-time feedback from the tactile fingertips (Figure~\ref{fig_touch_follow}). A feedback controller based on stabilising the tactile deformation (estimated from the SSIM and equivalent to contact force) dynamically adjusts the finger posture via activating the agonist and antagonist motors to ensure stable contact. During dynamic contact interactions, continuous variations in contact force and direction between the indenter and tactile fingertip induce changes in the tactile images, from which the Tactile SoftHand-A adaptively regulates its fingertip positions to preserve a preset contact threshold. This demonstrates the advantage of tactile sensing in counteracting contact disturbances. 

The second experiment demonstrates the tactile feedback mechanism in the Tactile SoftHand-A (Figure~\ref{fig_hand_control}) to synchronize gestures with a human hand and respond to external stimuli. The synchronization phase demonstrates effective mirroring of human hand movements, with a minimal delay attributed to the antagonistic tendon mechanism, which reduces reliance on passive elements within the finger joints, thereby enhancing responsiveness. Furthermore, the contact and slip detection phases demonstrate the Tactile SoftHand-A's ability to maintain a stable grasp while dynamically adjusting its gesture to prevent slippage (Figure~\ref{fig_hand_tac_grasping_results}). 

Finally, the third experiment comparatively analyzes the performance of tactile slip detection and grasp gesture compensation mechanisms using the teleoperation of the second experiment (Figure.~\ref{fig_hand_tac_stable_grasping}). Under disturbance-free conditions, both tactile-enabled and tactile-deprived grasping modalities achieve stable object retention. However, when dynamic disturbances were applied to the manipulated object, the non-tactile system fails to detect grasp failure (slip events), ultimately resulting in object loss. In contrast, the tactile-enabled Tactile SoftHand-A successfully demonstrates detecting slip occurrences and triggers compensatory grasp adjustments, thereby maintaining a stable grasp. This closed-loop tactile feedback control is crucial for tasks that require a secure grip, such as holding delicate objects or performing precision tasks. By effectively detecting and responding to slippage, the robotic hand can adjust to dynamic environmental conditions, significantly improving its functionality, reliability, and overall performance in real-world applications.

Overall, these three experiments using tactile feedback confirm the critical role of tactile perception in enhancing disturbance resistance and improving grasping reliability within unstructured environments, and how this complements the adaptivity and controllability of the SoftHand-A due to its antagonistic tendons and synergistic mechanism.

\section{Conclusion}
This paper shows the sophisticated control and adaptivity possible with a 3D-printed anthropomorphic SoftHand with an antagonist tendon mechanism and integrated tactile sensing. The coordinated control of agonist and antagonist motors allows for precise manipulation of finger joints, while the differential mechanism enhances grasping adaptivity. The tactile feedback system further improves the hand's responsiveness and stability in autonomous and teleoperation scenarios. These findings have implications for the application of tendon-driven dexterous hands, for example as prosthetics for assisting individuals with disabilities to perform daily grasping tasks and for remote operations in hazardous environments or precision surgeries to enable more accurate and reliable human-machine interactions. Future work could focus on refining these control mechanisms and exploring their applications in dexterous manipulation. Further, because the hand is 3D-printed, different combinations of agonist and antagonistic mechanisms coupled with the degree of actuation can be explored. This opens up a rich research area of adding greater dexterity to minimally-actuated anthropomorphic robot hands, so that it becomes possible to investigate systematically how to develop low-cost hands with human-like dexterity. To this end, we openly release all designs and fabrication instructions for creating the Tactile SoftHand-A as an enabler for others to build upon this work.  

\section{Funding}
CL, HL and YL were supported by the China Scholarship Council (CSC) and Bristol joint scholarship. EP and NL were supported by the Horizon Europe research and innovation program under grant agreement No. 101120823 (`MANiBOT: Advancing the physical intelligence and performance of roBOTs towards human-like bi-manual objects MANipulation') and a Royal Society International Collaboration Award (South Korea). NL was also supported by the Advanced Research + Invention Agency (`Democratising Hardware and Control For Robot Dexterity').

\bibliographystyle{SageH}
\bibliography{references.bib}

\begin{figure*}[t!]
		\centering
		\begin{tabular}[b]{c}
			\hspace{0cm}\includegraphics[width=1\textwidth,trim={10 0 0 0},clip]{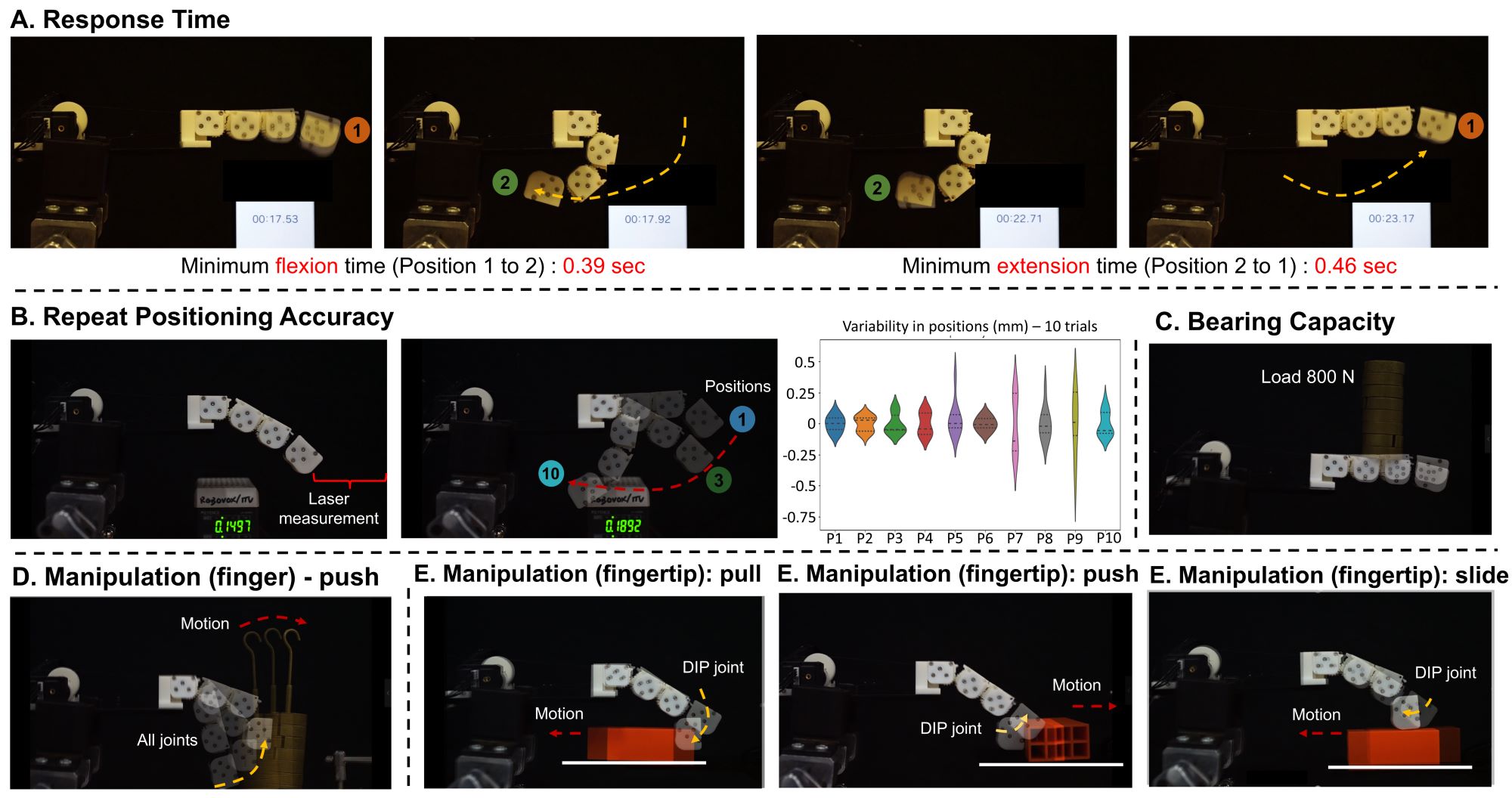} \\
		\end{tabular}
             \caption{Experimental evaluation of D-Type finger performance. A) {\em Response Time Test}: Evaluates the response time of the finger during flexion and extension movements between fully open or closed finger positions. B) {\em Repeat Positioning Accuracy Test}: Assesses the finger’s ability to return to predefined positions with precision. The laser rangefinder tracks the fingertip as it moves through 10 distinct positions over 10 trials. C) {\em Bearing Capacity Test}: Evaluates the finger’s load-bearing capability under static conditions. An 8\,N weight is applied to the fingertip and the drop measured under full loading of both motors. D) {\em Obstacle Resistance Test}: Evaluates the ability of the finger to displace a stationary obstacle in the path of motion. E) {\em Finger Manipulation Test}: Assesses the finger's ability to manipulate objects through controlled movements, pushing objects inward, outward and complex dexterous manipulations. }
		\label{fig_finger_supplement}%
		\vspace{4em}

		\centering
		\begin{tabular}[b]{c}
	\hspace{0cm}\includegraphics[width=1\textwidth,trim={15 0 0 0},clip]{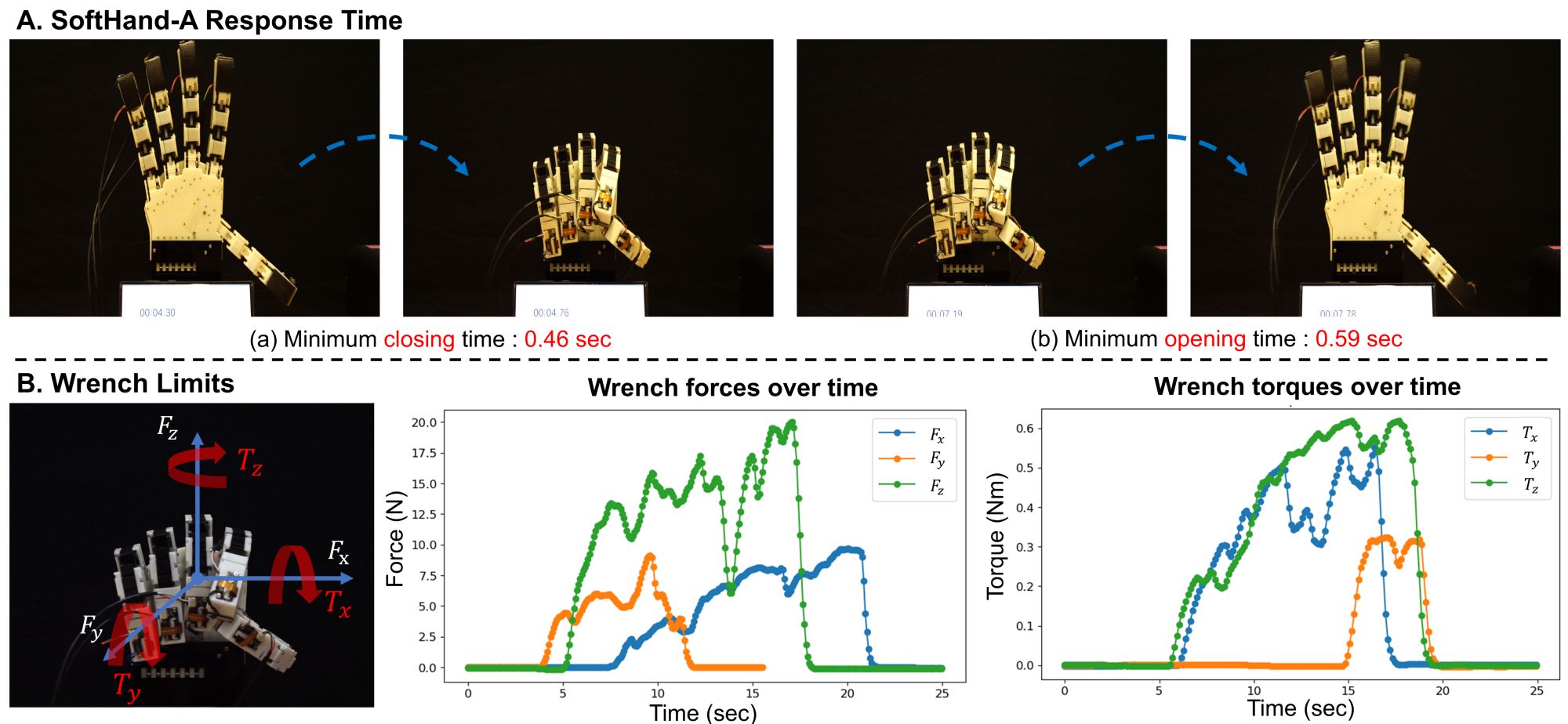} \\
		\end{tabular}
             \caption{Experimental evaluation of whole-hand performance. A) to the left, the minimum closing time from the open gesture to the closed gesture; to the right, the minimum opening time from the closed gesture to the open gesture. B) Wrench limits of Tactile SoftHand-A. To the left, the directions of forces and torques acting on the SoftHand-A, then the other two panels present the wrench forces and torques over time, respectively. These were obtained using force/torque sensors to measure the disturbances of the robotic hand under stable grasping conditions.}
		\label{fig_hand_response_wrench}%
		\vspace{0em}
\end{figure*}

\begin{figure*}[t!]
		\centering
		\begin{tabular}[b]{c}
	\hspace{0cm}\includegraphics[width=\textwidth,trim={0 0 0 0},clip]{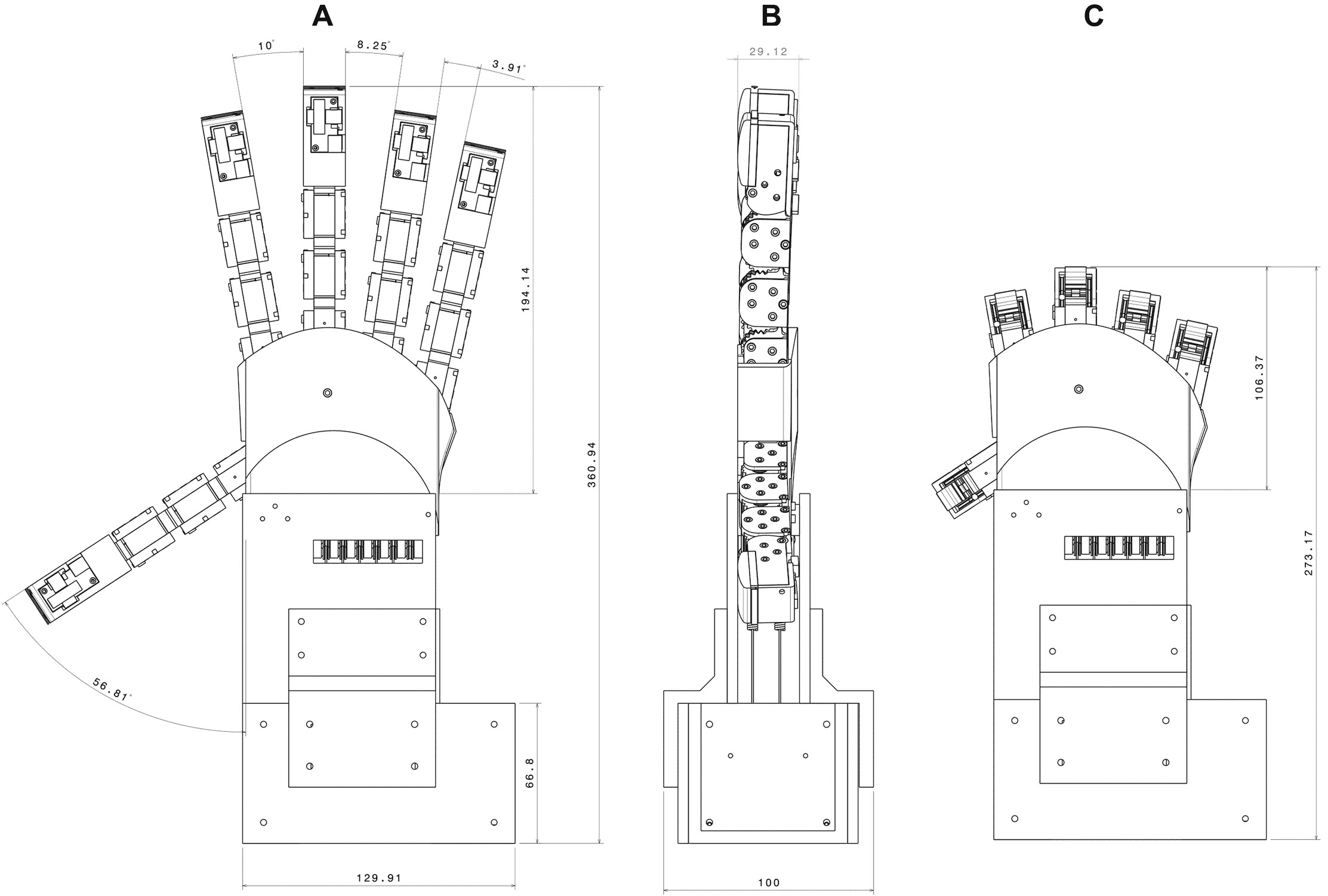} \\
		\end{tabular}
             \caption{Technical drawing of the Tactile SoftHand-A showing the components and their dimensions. A) Back view of the open hand. B) Left view of the open hand. C) Back view of the closed hand.}
		\label{appendix_fig_hand_size}%
		\vspace{8em}
		\centering
		\begin{tabular}[b]{c}
	\hspace{0cm}\includegraphics[width=\textwidth,trim={0 0 0 0},clip]{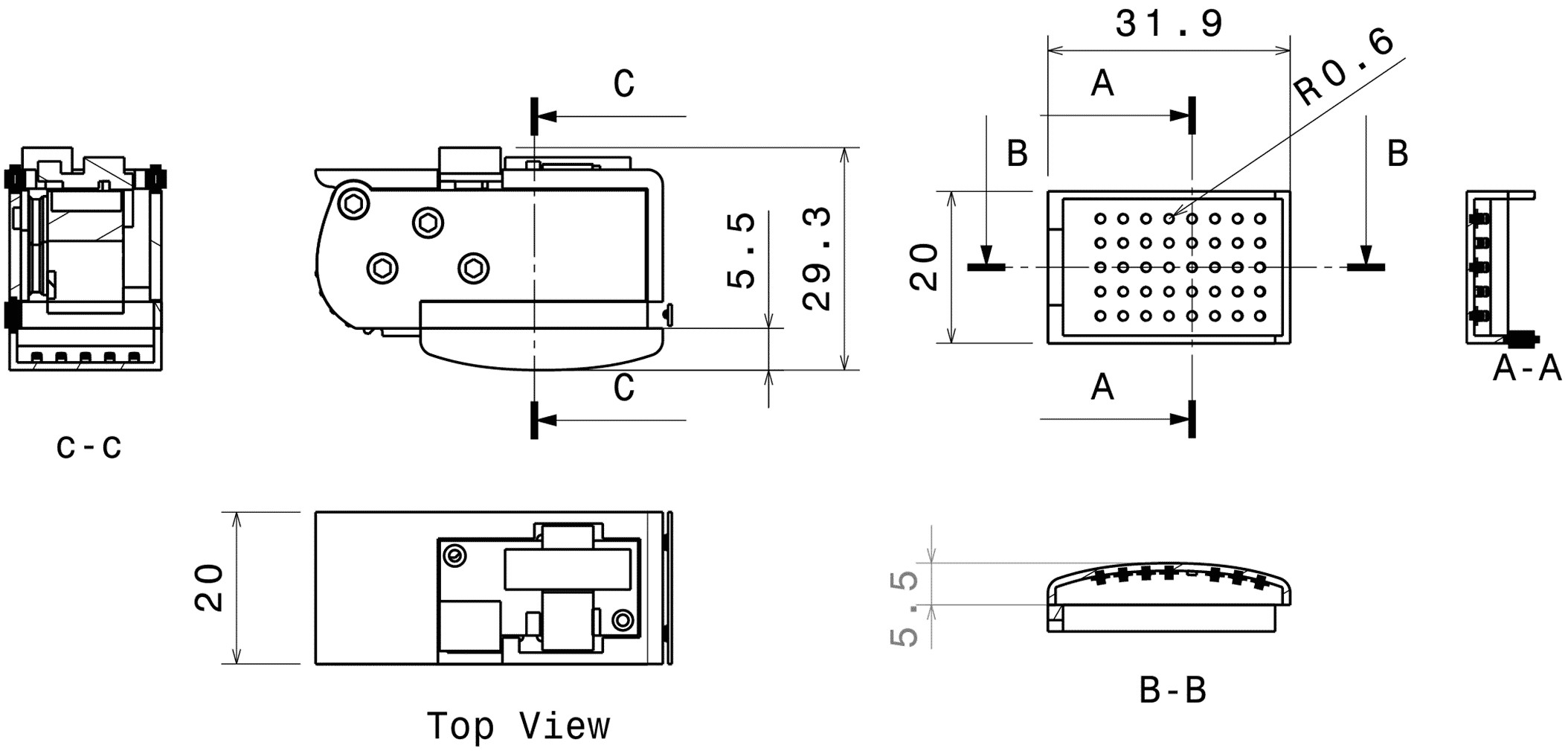} \\
		\end{tabular}
             \caption{Technical drawing of the tactile fingertip showing all components and their dimensions.}
		\label{appendix_tac_fingertip_size}%
		\vspace{0em}
\end{figure*}

\section{Appendix}

\subsection{Single Finger Testing}

The single tests reported in Table~\ref{tab:1} are shown in Figure~\ref{fig_finger_supplement}.

\subsection{Whole Hand Testing}

The whole hand tests reported in Table~\ref{tab:2} are shown in Figure~\ref{fig_hand_response_wrench}

\subsection{Technical Diagrams and Details}

A technical drawing of the Tactile SoftHand-A is given in Figure~\ref{appendix_fig_hand_size} and a technical drawing of the tactile fingertip given in Figure~\ref{appendix_tac_fingertip_size}. The hand is actuated by two HIWONDER HX-35H motors. At 11.1 volts, these motors provide a torque of 25\,kg/cm. The hand is manufactured using multi-material 3D printing (Stratasys J826). The hand’s skeleton and most of the opaque solid parts are made from Verowhite material. The parts resembling acrylic plates are made from Veroclear. Transparent soft materials and the black tactile epidermis use Agilus clear and Agilus black. Interfacing is via UART serial communication.The entire tactile fingertip is produced in a single print, incorporating white hard phalanges, black soft tactile skin, transparent white filler, white markers, and a transparent hard window. Each fingertip camera module is connected to the host computer via a dedicated COM port. 

\subsection{Control diagram}

\begin{figure}[h]
    \centering
    \includegraphics[width=1\linewidth]{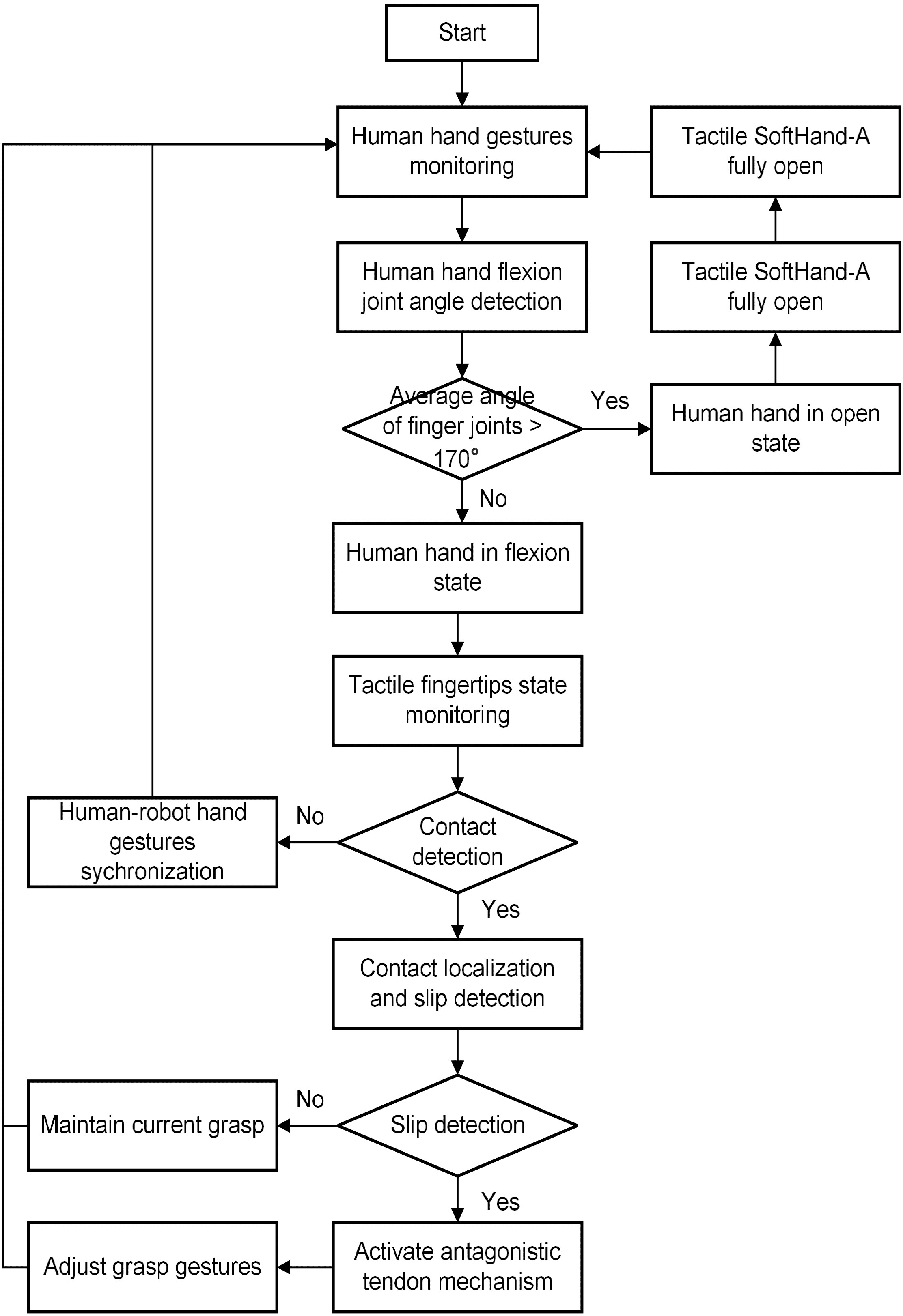}
    \caption{Flow chart depicting the process for grasp control under tactile feedback.}
    \label{fig_tactile_flow_chart}
\end{figure}

A control flow diagram is given in Figure~\ref{fig_tactile_flow_chart}.

\end{document}